\documentclass{article}

\usepackage[utf8]{inputenc}
\usepackage[T1]{fontenc}
\usepackage{lmodern}
\usepackage[letterpaper,
            left=1.5in, right=1in,
            top=1in, bottom=1in,
            textwidth=5.5in]{geometry}

\usepackage{amsmath}
\usepackage{amsfonts}
\usepackage{amssymb}
\usepackage{nicefrac}
\usepackage{booktabs}
\usepackage{array}
\usepackage{tabularx}
\usepackage{multirow}
\usepackage{graphicx}
\usepackage{xcolor}
\usepackage{microtype}
\usepackage{quoting}
\usepackage{lineno}
\usepackage{CJKutf8}
\usepackage[hidelinks,colorlinks=true,linkcolor=black,citecolor=blue,urlcolor=blue]{hyperref}
\usepackage{url}

\makeatletter
\renewcommand{\maketitle}{%
  \begin{center}
    \rule{\textwidth}{4pt}\\[0.25in]
    {\fontsize{17}{20}\selectfont\bfseries \@title \par}
    \vspace{0.25in}
    \rule{\textwidth}{1pt}\\[0.5em]
    {\large \@author \par}
    \vspace{0.5em}
    {\@date}
  \end{center}
  \vspace{1em}
}
\makeatother

\newenvironment{modelout}{%
  \begin{quoting}[indentfirst=false,leftmargin=1em,rightmargin=1em,vskip=0.3em]%
  \small\itshape
}{%
  \end{quoting}
}

\title{LLiMba: Sardinian on a Single GPU --- Adapting a 3B Language Model to a Vanishing Romance Language}

\author{%
  Luca Ballore \\
  \texttt{luca.ballore@gmail.com}
}

\date{}

\begin{document}

\nolinenumbers

\maketitle

\begin{abstract}
Sardinian, a Romance language with roughly one million speakers, has minimal presence in modern NLP. Commercial services do not support it, and current language models do not produce it reliably. We present LLiMba, a 3B parameter Sardinian-ready model adapted from Qwen2.5-3B-Instruct through continued pretraining (CPT) and supervised fine-tuning (SFT) on a single 24~GB consumer GPU. The corpus contains 11.5 million tokens of Sardinian spanning LSC, Logudorese, and Campidanese, augmented with 2.4 million tokens of related Romance text as replay against register blurring. After CPT the model reaches a perplexity of 6.76 on held out Sardinian and outperforms the base across all six FLORES-200 directions. We compare five SFT configurations under matched conditions: full fine-tuning, LoRA r64, rsLoRA r128, rsLoRA r256, and DoRA r256. rsLoRA r256 wins on every direction into Sardinian, reaching 28.5 BLEU from English against 17.3 after CPT and 21.0 with full fine-tuning. The rank ablation places r128 between LoRA r64 and rsLoRA r256 on BLEU but reveals failure modes invisible to the metric, including leakage across scripts no other variant produces. LoRA r64 retains less factual content from SFT than configurations at higher rank and produces more confident fabrications, though all methods fabricate on content absent from training. DoRA r256 yields the smallest gap between training and evaluation but the worst factual accuracy. The findings indicate that adapter capacity matters more than the choice among LoRA variants for adapting a Romance pretrained base to a low resource Romance target, that stronger regularization is not uniformly beneficial, and that translation metrics smoothly order configurations whose qualitative behavior differs categorically. Perplexity comparisons across scripts must account for byte fallback tokenization, which deflates the metric for scripts other than Latin.
\end{abstract}

\section{Introduction}
\label{sec:intro}

Sardinian (ISO 639-3: srd) is a Romance language spoken by roughly one million people on the island of Sardinia, Italy. UNESCO classifies it as endangered. Despite the language's demographic depth and active community of writers, it has effectively no presence in commercial NLP infrastructure. No major translation service supports it, no voice assistants understand it. Commercial large language models, when prompted in Sardinian, default to Italian, Portuguese, Spanish, Catalan, French, or even English; some refuse the prompt entirely. One model we tested confused the autonym \emph{sardu} with the fish (\emph{sardine}).

The causes of this invisibility are structural. Proprietary models do not target Sardinian because the user base is too small to justify data acquisition costs. Open models do not target it because the available training data is too sparse to register in web-scale corpora. Standard low-resource NLP datasets (OSCAR, the Leipzig Corpora Collection, the eBible corpus) contain little or no usable Sardinian text. The Sardinian Wikipedia exists but is smaller than other Romance Wikipedias by an order of magnitude or more.

The phylogenetic position of Sardinian suggests that adaptation should be tractable. Sardinian shares Latin etymology, Romance morphology, and SVO syntax with the broader Romance family, and is particularly close to Italian, Spanish, Portuguese, and Catalan. A multilingual base model that encodes those structures already has most of the linguistic scaffolding required. The adaptation task reduces to teaching the language-specific lexicon, orthography, and idiom rather than learning a language family from scratch. This makes Sardinian a useful case for studying minimal-data adaptation within an already-understood language family.

The closest methodological reference is Chen et al.~(2025)~\cite{chen2025}, who report a two-stage continued-pretraining and supervised-fine-tuning pipeline for Tibetan, also based on Qwen2.5-3B. Their setting is informative but distant from ours in three respects. First, Tibetan is written in a non-Latin syllabary, which interacts with byte-fallback tokenization in ways that can artificially deflate perplexity. Second, Tibetan is typologically distant from every language in Qwen's training data, so adaptation begins from minimal prior. Third, Chen et al.\ report SFT-stage BLEU under 1 on their best translation direction, leaving open how a CPT and SFT pipeline performs when the base model already has substantial prior support for the target language family.

We present \emph{LLiMba}, an open Sardinian-capable language model adapted from Qwen2.5-3B-Instruct on a single 24~GB consumer GPU. The training pipeline collects approximately 13.5 million tokens of Sardinian from heterogeneous sources, retains roughly 11.5 million tokens after deduplication and language filtering, augments them with approximately 2.4 million tokens of related Romance text used as replay against catastrophic forgetting, then applies continued pretraining followed by supervised fine-tuning. Beyond producing the model itself, the work contributes an empirical comparison of five supervised fine-tuning configurations under matched data, hardware, and evaluation conditions:

\begin{itemize}
  \item full fine-tuning,
  \item LoRA at rank 64,
  \item rsLoRA at rank 256, and
  \item DoRA at rank 256.
\end{itemize}

Alongside the quantitative comparison, we document failure modes specific to low-resource Romance adaptation: translation calques that survive iterative data cleaning, prompt-phrasing sensitivity for factual recall, and high-temperature compositional artifacts that appear in no training example. Our perplexity figures further illustrate why such measurements must be compared with care across languages, since byte-fallback tokenization can artificially compress loss values for non-Latin scripts. The full pipeline runs on hardware available to individual researchers and small labs, with configurations and results documented for reproduction.

\section{Background}
\label{sec:background}

The most directly comparable line of work is the Tibetan adaptation literature. Chen et al.~(2025)~\cite{chen2025} propose a two-stage continued pretraining and supervised fine-tuning pipeline on Qwen2.5-3B, the same base model and pipeline structure we adopt. T-LLaMA (Lv et al.\ 2025)~\cite{lv2025} adapted LLaMA2-7B to Tibetan through continued pretraining with vocabulary expansion on a 2.2-billion-character corpus. Banzhida (Pan et al.\ 2025)~\cite{pan2025} subsequently scaled the approach, continuing to pretrain Qwen2.5-7B on a curated Tibetan dataset alongside Chinese and English replay data. Together these works establish that two-stage adaptation is workable for low-resource languages, but they operate on Tibetan, which differs from Sardinian in two ways central to our analysis: it is typologically and orthographically distant from anything in the multilingual base model's pretraining distribution, and the Tibetan script interacts with byte-fallback tokenization in ways that complicate loss-based metrics. The base model in our work, Qwen2.5-3B-Instruct (Yang et al.\ 2024)~\cite{yang2024}, includes substantial Romance-language pretraining, which changes both the starting point of adaptation and the kinds of failure modes that emerge.

Hu et al.~(2021)~\cite{hu2021} introduced LoRA as a parameter-efficient alternative to full fine-tuning, training low-rank adapters in place of full weight updates. Kalajdzievski~(2023)~\cite{kalajdzievski2023} showed that LoRA's conventional $\alpha/r$ scaling factor causes gradient collapse at higher ranks; the rsLoRA correction ($\alpha/\sqrt{r}$) restores stability and makes higher ranks practical. DoRA (Liu et al.\ 2024)~\cite{liu2024} decomposes weight updates into magnitude and direction and adapts them separately, aiming to preserve the base model's directional structure during fine-tuning. Biderman et al.~(2024)~\cite{biderman2024} compared LoRA and full fine-tuning across continued pretraining and instruction tuning on programming and mathematics, reporting that LoRA underperforms full fine-tuning when the target domain is far from the pretraining distribution while better preserving capabilities outside the target domain. That finding directly informed our decision to use full fine-tuning for the CPT stage, where the language-adaptation domain shift is largest, and to compare adapter variants only at the SFT stage where the shift is smaller. Baqar and Khanda~(2025)~\cite{baqar2025} compared RAG, LoRA, and DoRA on factuality across 20{,}000 FAQ queries and report that adapter methods can produce fluent output that fails to ground in the training data, a trade-off between fluency and factual grounding. Our SFT comparison reproduces this pattern in the low-resource adaptation setting, where the magnitude of the effect varies systematically across LoRA, rsLoRA, and DoRA at matched rank.

For evaluation, the FLORES-200 benchmark (NLLB Team 2022)~\cite{nllb2022} provides parallel sentences across 200 languages, including Sardinian, and is widely used for low-resource translation. We adopt it for our six-direction translation comparison and run all evaluations through \texttt{lm-evaluation-harness} (Gao et al.\ 2023)~\cite{gao2023} for consistency across model variants. We report both BLEU and chrF; chrF is more robust to the morphological richness and dialectal variation present in Sardinian, where valid synonyms or alternative forms are penalized by exact-match BLEU.

\section{Data}
\label{sec:data}

The training data falls into three groups: a Sardinian pretraining corpus augmented with related Romance replay text, a supervised fine-tuning dataset built from instruction pairs, and a held-out evaluation set drawn from FLORES-200.

\subsection{Pretraining corpus}
\label{sec:data-cpt}

We collected approximately 13.5 million tokens of Sardinian text from heterogeneous sources. After deduplication and language filtering, around 11.5 million tokens of Sardinian remain in the training corpus. Table~\ref{tab:cpt-corpus} lists the composition after preparation.

\begin{table}[h]
  \centering
  \caption{Pretraining corpus composition after preparation.}
  \label{tab:cpt-corpus}
  \begin{tabular}{lrr}
    \toprule
    Source & Documents & Tokens \\
    \midrule
    Web scrape (six sites)                  & 8{,}110  & 4.90M \\
    Sardinian Wikipedia                     & 6{,}309  & 2.58M \\
    GlotCC CommonCrawl                      & 2{,}270  & 1.77M \\
    Translated books (PDF, EPUB, markdown)  & 409      & 2.01M \\
    Poetry anthologies                      & 436      & 176K  \\
    Bilingual text and song lyrics          & 84       & 39K   \\
    \midrule
    \textbf{Total Sardinian}                & \textbf{17{,}618} & \textbf{11.48M} \\
    Romance replay (Wikipedia)              & 652      & 2.44M \\
    \midrule
    \textbf{Combined corpus}                & \textbf{18{,}270} & \textbf{13.93M} \\
    \bottomrule
  \end{tabular}
\end{table}

The web scrape covers six verified-live sites publishing in Sardinian on news, culture, technology, and provincial institutional topics. The book material consists of professional Sardinian translations of world literature; this provides the corpus's most extended literary prose, with stylistic and lexical variety that web sources alone cannot match. Poetry anthologies cover regional verse from 1400 to 1900, with line breaks preserved during extraction to retain poetic structure. GlotCC contributes filtered CommonCrawl text and overlaps substantially with the web scrape; the overlap is removed during preparation, reducing GlotCC from 3{,}790 raw documents to 2{,}270 in the final corpus.

The corpus deliberately spans the three main written variants: LSC (\emph{Limba Sarda Comuna}, the standardized form), Logudorese, and Campidanese. This reflects how Sardinian is actually published: news sites use LSC, institutional documents use Campidanese, and literary works span all three. The model targets LSC for output but is exposed to all variants on input.

Approximately 2.4 million tokens of related Romance text, drawn from the Italian, Spanish, Portuguese, and Catalan Wikipedias, are mixed into the corpus to mitigate catastrophic forgetting and prevent representational blurring between Sardinian and Italian. Italian dominates the replay, with smaller shares of Spanish, Portuguese, and Catalan. The replay text carries no language tag; the model learns to distinguish languages from the text itself, matching the conditions it will face at inference time.

The corpus required substantial cleaning. Sardinian web sources commonly mix Sardinian body text with Italian navigation, headers, and footers. Standard language detection tools do not recognize Sardinian and classify it variably as Italian, Portuguese, Spanish, or Catalan; we exploit this rather than fight it, retaining documents that classify as any of those four and removing only documents flagged as English, German, or French. Online dictionary content with highly repetitive template structures was extracted from pretraining text to avoid overfitting on its surface form. The author, a native speaker, reviewed approximately 150 documents to spot-check quality across sources; the review confirmed that mixed-language documents and book attribution lines were worth retaining for the value of their Sardinian content.

After document chunking with overlap, the corpus yields 19{,}152 training examples.

\subsection{SFT data}
\label{sec:data-sft}

The SFT pool combines machine-translated instruction data with native-curated material across four buckets, summarized in Table~\ref{tab:sft-data}.

\begin{table}[h]
  \centering
  \caption{SFT data composition before deduplication and upsampling.}
  \label{tab:sft-data}
  \begin{tabular}{lrr}
    \toprule
    Source & Pairs & Tokens \\
    \midrule
    Capybara translated to Sardinian                 & 10{,}517 & 12.68M \\
    Translation pairs                                & 2{,}020  & 206K   \\
    Synthesized instructions (Claude with native review) & 448  & 77K    \\
    Song-related pairs                               & 142      & 36K    \\
    \midrule
    \textbf{Total raw}                               & \textbf{13{,}127} & \textbf{13.00M} \\
    \bottomrule
  \end{tabular}
\end{table}

The bulk comes from the Capybara dataset (LDJnr)~\cite{ldjnr2023}, a multi-turn instruction tuning collection, machine-translated into Sardinian using NLLB-200 3.3B (NLLB Team 2022)~\cite{nllb2022}, itself a model that runs on the same consumer hardware as our training pipeline. Capybara provides diversity across instruction types (literature, mathematics, science, reasoning, conversation). Translation quality is uneven; the output was cleaned through a combination of automated heuristics (filtering very short and very long responses, dropping entries whose Sardinian-side text failed basic checks) and native-speaker review. The translated pool nevertheless contains residual calques, Italian-shaped grammatical structures rendered with Sardinian vocabulary, that survive iterative cleaning. We treat these as a known limitation and return to them in Section~\ref{sec:limitations}.

The translation pairs collect parallel sentences across multiple sources, providing explicit supervision for translation tasks. Synthesized instructions were generated with the assistance of Anthropic's Claude using Sardinian grammar references as anchoring context, and reviewed entry-by-entry by the author. Song-related pairs cover retrieval, identification, and content questions about Sardinian song lyrics.

After deduplication, 12{,}716 pairs remain. The synthesized bucket contributes 422 of these (the remaining 26 of the original 448 were duplicates of other entries and were removed during deduplication). The synthesized bucket is then upsampled by a factor of five during dataset assembly, reflecting the higher native-review confidence of those pairs relative to the machine-translated and bulk translation pools, contributing 1{,}688 additional copies. The final SFT pool contains 14{,}404 pairs and approximately 12.8 million tokens.

SFT examples vary in their system prompt configuration. The majority carry a Sardinian language system prompt that frames the assistant as a Sardinian language helper, with translation examples using prompts that name the target language explicitly. Approximately six percent of the pool (around 875 of the 14,404 final pairs) carries a system prompt written in another language: roughly 300 in Italian, 250 in English, 150 in Spanish, 100 in Portuguese, and 75 in French. A smaller share carries no system prompt at all. The variation is intentional, intended to teach the model that the language of the response should track the user's request rather than the language of the system prompt, and to expose the model to realistic deployment conditions where a developer might wrap the model in a non-Sardinian system instruction, or none at all, while still expecting Sardinian output.

\subsection{Evaluation data}
\label{sec:data-eval}

Translation evaluation uses 997 parallel sentences from FLORES-200, aligned across Sardinian, Italian, English, Spanish, French, and Portuguese. We report results on six translation directions: English-to-Sardinian, Italian-to-Sardinian, Spanish-to-Sardinian, Sardinian-to-English, Sardinian-to-Italian, and Sardinian-to-Spanish.

We additionally maintain a qualitative probe set of Sardinian prompts spanning conversation, translation, factual question-answering on Sardinian culture and history, text continuation, creative writing, and grammatical analysis. The set is run uniformly across all model variants for native-speaker comparison. The probes complement the BLEU and chrF figures and surface failure modes that translation metrics cannot capture.

\section{Method}
\label{sec:method}

We adapt Qwen2.5-3B-Instruct (Yang et al.\ 2024)~\cite{yang2024} to Sardinian in two stages following the structure of Chen et al.~(2025)~\cite{chen2025}. Stage~1 is continued pretraining (CPT) on the Sardinian corpus from Section~\ref{sec:data-cpt}, applied as full fine-tuning in bfloat16. Stage~2 is supervised fine-tuning (SFT) on the instruction data from Section~\ref{sec:data-sft}, run independently in five configurations to compare full fine-tuning against four adapter variants. Both stages run on a single NVIDIA RTX~4090 with 24~GB of VRAM, using the HuggingFace \texttt{transformers}, \texttt{peft}, and \texttt{trl} libraries. All random seeds are fixed at 42.

We use Qwen2.5-3B-Instruct rather than the base model because this variant retains a usable instruction following scaffold that CPT partially erases and that SFT can then re-anchor with a small instruction dataset; starting from the base model would require a substantially larger SFT pool to teach instruction following from scratch. The 3B parameter size fits in 24~GB at bfloat16 with gradient checkpointing and 8-bit optimizer states, which makes full fine-tuning feasible on the available hardware. Qwen2.5 was chosen over alternatives at this size for its multilingual pretraining, which spans the major Romance languages and gives the model a useful prior for Sardinian.

\subsection{Continued pretraining}
\label{sec:method-cpt}

CPT updates all parameters of the base model using the configuration in Table~\ref{tab:cpt-config}.

\begin{table}[h]
  \centering
  \caption{Continued pretraining configuration.}
  \label{tab:cpt-config}
  \begin{tabular}{ll}
    \toprule
    Hyperparameter & Value \\
    \midrule
    Precision                 & bfloat16 \\
    Attention                 & Flash Attention 2 \\
    Optimizer                 & paged AdamW 8-bit \\
    Learning rate             & 5e-5 \\
    Schedule                  & cosine, 50 warmup steps \\
    Effective batch size      & 16 (batch 1, 16 gradient accumulation steps) \\
    Sequence length           & 4096 \\
    Sequence packing          & disabled \\
    Gradient checkpointing    & enabled \\
    Epochs                    & 2 \\
    Held-out evaluation split & 2.5\% \\
    \bottomrule
  \end{tabular}
\end{table}

We use full fine-tuning for this stage rather than a parameter efficient adapter based on Biderman et al.~(2024)~\cite{biderman2024}, who report that LoRA underperforms full fine-tuning when the target domain is far from the pretraining distribution. Teaching a new language qualifies as a large domain shift, and the language modeling pressure is best applied to the full parameter set rather than a low-rank subspace.

We disable sequence packing despite its throughput benefits because packing allows attention to leak across document boundaries within a packed sequence. On a heterogeneous corpus where short Wikipedia stubs sit alongside long book chapters, this cross-contamination homogenizes representations across genres and, in our preliminary runs, produced markedly degraded model quality at matched hyperparameters. We treat unpacked training as required rather than optional. Long documents occasionally exceed the 4096 sequence length, so we chunk such documents into overlapping windows with a 128-token overlap rather than truncate them. The overlap preserves local context across chunk boundaries.

Romance replay text is interleaved with Sardinian text by random shuffling at the document level. The model receives no language tag and learns to distinguish languages from the text itself, matching inference-time conditions.

Fitting full fine-tuning of a 3B parameter model into 24~GB of VRAM is the core hardware constraint of this work. The decisive component is the paged 8-bit AdamW optimizer (Dettmers et al.\ 2023)~\cite{dettmers2023}: a standard fp32 AdamW would require roughly 24~GB for optimizer states alone, exhausting the memory budget before weights or activations are allocated. The 8-bit variant reduces these states approximately fourfold, and the paged mechanism allows transient spills to system memory under pressure rather than triggering OOM failures. Combined with bfloat16 weights, gradient checkpointing for activations, and an effective batch of 16 reached through gradient accumulation rather than per-device batching, peak VRAM use lands at 22 to 23~GB, leaving a usable margin. Without the 8-bit optimizer, full fine-tuning of a 3B model on this hardware would not be feasible.

The CPT run completes in approximately 5.5 hours of wall-clock time on the single RTX~4090.

\subsection{Supervised fine-tuning}
\label{sec:method-sft}

We compare five SFT configurations starting from the same CPT-adapted model:

\begin{itemize}
  \item \textbf{Full fine-tuning}: all parameters updated in bfloat16, mirroring the SFT recipe of Chen et al.~(2025)~\cite{chen2025} with a small learning rate.
  \item \textbf{LoRA r64}: low-rank adapters at rank 64 with $\alpha=128$, a conventional configuration in the style of Hu et al.~(2021)~\cite{hu2021}.
  \item \textbf{rsLoRA r128}: rank-stabilized LoRA at rank 128 with $\alpha=128$, included as a rank ablation between the conventional LoRA configuration and the higher-rank rsLoRA run.
  \item \textbf{rsLoRA r256}: rank-stabilized LoRA at rank 256 with $\alpha=256$ and the $\alpha/\sqrt{r}$ scaling correction of Kalajdzievski~(2023)~\cite{kalajdzievski2023}. The square-root scaling restores stable gradient norms at higher ranks and makes rank 256 trainable in practice.
  \item \textbf{DoRA r256}: weight-decomposed LoRA at rank 256 with $\alpha=256$ (Liu et al.\ 2024)~\cite{liu2024}. DoRA decomposes each weight update into magnitude and direction components and adapts them separately, with the goal of preserving the base model's directional structure.
\end{itemize}

All five configurations train on the SFT pool from Section~\ref{sec:data-sft} using the ChatML chat template, with the loss restricted to assistant completions so that system prompts and user turns are masked out. Hyperparameters for each method appear in Table~\ref{tab:sft-config}.

\begin{table}[h]
  \centering
  \caption{SFT method configurations. All five train for 2 epochs on the same data with completion-only loss, sequence length 4096, effective batch size 16 (batch 1, 16 gradient accumulation steps), 50 warmup steps, a 5\% held-out evaluation split, and Flash Attention 2. The four adapter configurations use LoRA dropout 0.05 and target the q, k, v, o, gate, up, and down projection matrices.}

  \label{tab:sft-config}
  \begin{tabular}{lrr}
    \toprule
    Configuration    & Learning rate & $\alpha$ \\
    \midrule
    Full fine-tuning & 1e-5 & N/A \\
    LoRA r64         & 2e-4 & 128 \\
    rsLoRA r128      & 2e-5 & 128 \\
    rsLoRA r256      & 2e-5 & 256 \\
    DoRA r256        & 2e-5 & 256 \\
    \bottomrule
  \end{tabular}
\end{table}

The five runs are otherwise identical: same data, same chat template, same loss formulation, same effective batch size, same hardware. This isolates the adapter method and rank as the variables across runs. The rsLoRA r256 and DoRA r256 runs use rank 256 to match Biderman et al.'s observation that ranks at or above 256 are required to close the gap with full fine-tuning. LoRA r64 is included as the conventional reference point, and rsLoRA r128 as an intermediate rank that holds the LoRA family's $\alpha/\sqrt{r}$ scaling fixed while halving the adapter parameter count of rsLoRA r256.

The model selected for deployment is the rsLoRA r256 SFT model, on the strength of the results in Section~\ref{sec:results}. The other four are reported as comparison points and as evidence for the rank-versus-method-choice analysis in Section~\ref{sec:discussion}.

\section{Results}
\label{sec:results}

\subsection{Translation benchmarks}
\label{sec:results-translation}

Table~\ref{tab:translation} reports BLEU and chrF for the base model, the CPT-adapted model, and the five SFT variants across six FLORES-200 translation directions. All evaluations use the same 997-sentence set, greedy decoding, and the template prompts described in Section~\ref{sec:data-eval}.

\begin{table}[h]
  \centering
  \caption{BLEU and chrF on FLORES-200 across six translation directions. Best score per direction in bold.}

  \label{tab:translation}
  \resizebox{\textwidth}{!}{%
  \begin{tabular}{lrrrrrrr}
    \toprule
    Direction & Base & CPT & Full FT & LoRA r64 & rsLoRA r128 & rsLoRA r256 & DoRA r256 \\
    \midrule
    EN-to-SC & 2.75 / 27.41  & 17.26 / 47.81 & 21.04 / 50.20 & 23.60 / 53.09 & 25.33 / 54.57 & \textbf{28.47 / 56.80} & 23.00 / 52.40 \\
    IT-to-SC & 2.16 / 27.52  & 12.71 / 44.83 & 16.45 / 48.01 & 18.52 / 50.13 & 19.64 / 51.00 & \textbf{21.25 / 52.08} & 17.70 / 49.19 \\
    ES-to-SC & 1.99 / 26.39  & 11.36 / 43.35 & 14.28 / 45.92 & 16.23 / 47.66 & 17.03 / 48.40 & \textbf{18.57} / \textbf{49.41} & 15.68 / 47.19 \\
    SC-to-EN & 11.73 / 44.55 & 33.52 / 62.78 & 38.05 / 62.67 & 39.61 / 63.59 & 40.69 / 64.23 & \textbf{41.28 / 64.64} & 38.98 / 63.10 \\
    SC-to-IT & 2.90 / 33.38  & 16.53 / \textbf{48.83} & \textbf{18.12} / 47.81 & 16.46 / 45.80 & 16.90 / 46.24 & 17.61 / 47.25 & 16.53 / 45.79 \\
    SC-to-ES & 5.67 / 36.98  & \textbf{19.31 / 47.76} & 19.08 / 47.05 & 18.62 / 46.35 & 18.83 / 46.62 & 18.57 / 46.27 & 18.84 / 46.54 \\
    \bottomrule
  \end{tabular}}
\end{table}

CPT delivers most of the translation gain. Across all six directions, the CPT-adapted model improves over the base by between four and seven times on BLEU. The largest absolute gain is on SC-to-EN, which moves from 11.73 to 33.52 BLEU; the largest relative gain is EN-to-SC, which moves from a near-zero 2.75 to 17.26. SFT then adds smaller increments on top, comparable in magnitude to the spread between SFT methods themselves and an order of magnitude smaller than the base-to-CPT step. CPT does the heavy lifting of teaching the model Sardinian, and SFT sharpens what is already there rather than introducing new capability.

rsLoRA at rank 256 is the strongest SFT configuration for translation into Sardinian. On the three into-Sardinian directions, rsLoRA r256 sits at the top by a clear margin: 28.47 BLEU on EN-to-SC against 23.60 for LoRA r64, 21.04 for full fine-tuning, and 23.00 for DoRA r256. The pattern repeats on IT-to-SC and ES-to-SC. rsLoRA applies the $\alpha/\sqrt{r}$ scaling correction of Kalajdzievski~(2023)~\cite{kalajdzievski2023}, motivated by analysis showing that the conventional $\alpha/r$ scaling becomes ineffective at high rank; Section~\ref{sec:discussion-capacity} unpacks the scaling math. SC-to-EN follows the same ordering with a smaller absolute spread, since the base model already had strong English generation; CPT supplied the Sardinian comprehension, and SFT contributed only a final polish.

Translation from Sardinian into the closer Romance languages saturates at CPT. SC-to-IT and SC-to-ES are essentially flat from CPT onward, with every SFT method underperforming or matching CPT on at least one of the two directions. CPT wins SC-to-ES (19.31 / 47.76) over all five SFT variants. Full fine-tuning narrowly wins SC-to-IT on BLEU (18.12 vs CPT's 16.53), but its chrF (47.81) is lower than CPT's (48.83). The SFT data evidently provides supervision for translation into Sardinian and for instruction following, but adds little new signal for translation from Sardinian into the closer Romance languages; the base model's existing Romance generation, combined with CPT's Sardinian reading comprehension, is already close to the ceiling that this corpus can support.

DoRA at rank 256 underperforms LoRA at rank 64 despite four times the capacity. DoRA r256 trails LoRA r64 on four of six directions: EN-to-SC (23.00 vs 23.60), IT-to-SC (17.70 vs 18.52), ES-to-SC (15.68 vs 16.23), and SC-to-EN (38.98 vs 39.61). DoRA holds a marginal edge on SC-to-IT (16.53 vs 16.46) and SC-to-ES (18.84 vs 18.62), both directions where the SFT methods cluster tightly and the absolute differences are within noise. DoRA's premise that adapters should preserve directional structure does not predict this outcome. The qualitative findings in Section~\ref{sec:results-qualitative} point the same way, and we discuss possible mechanisms in Section~\ref{sec:discussion}.

The rsLoRA rank ablation orders the three LoRA-family configurations smoothly. On every into-Sardinian direction, rsLoRA r128 sits between LoRA r64 and rsLoRA r256 by margins that exceed the bootstrap standard error. EN-to-SC moves from 23.60 to 25.33 to 28.47 BLEU, IT-to-SC from 18.52 to 19.64 to 21.25, ES-to-SC from 16.23 to 17.03 to 18.57, SC-to-EN from 39.61 to 40.69 to 41.28. The two from-Sardinian directions (SC-to-IT, SC-to-ES) show the same ordering at smaller absolute spreads, consistent with the saturation behavior already noted for those directions. Translation metrics order the three rsLoRA-family configurations cleanly by rank, and the smooth metric ordering does not reflect the qualitative gap that Section~\ref{sec:results-qualitative} documents.

\subsection{Qualitative findings}
\label{sec:results-qualitative}

The translation table shows that rsLoRA r256 dominates into Sardinian and that DoRA r256 underperforms despite higher rank, but it does not show why. The qualitative probes reveal additional axes along which the SFT methods diverge. All variants produce fluent Sardinian after CPT; the methods differ sharply in whether the factual content of that fluent Sardinian is grounded in the training data, and in whether the words it uses are attested Sardinian vocabulary. Three probes illustrate the patterns most clearly.

The first asks ``Chie fiat Gigi Riva?'' (Who was Gigi Riva?). Riva (1944--2024) was an Italian footballer who joined Cagliari Calcio in 1963 and led the club to its only Serie~A title in the 1969--70 season, scoring 35 goals in 42 appearances for the Italian national team and earning the nickname ``Rombo di Tuono'' from the journalist Gianni Brera. He is a major figure in Sardinian sporting memory and is covered in our pretraining corpus through Wikipedia and web sources. The base model misclassifies him as ``unu actor sardu'' (a Sardinian actor). CPT correctly identifies him as a famous Italian footballer but offers no further detail. Full fine-tuning adds that he was Cagliari's best striker, still vague but accurate. LoRA r64 invents biographical specifics with confidence, claiming he was nicknamed ``su Messi sardu'' (the Sardinian Messi), born in Tergu, won the 1982 World Cup in Scotland, and won the Intercontinental Cup with three goals. None of these claims is true. DoRA r256 is even worse: it places his birth in Casteddu (Cagliari) in 1947 (Leggiuno, 1944), claims he played four seasons for Juventus and five for Milan (he played for Cagliari, not either club), gives him the nickname ``Su Gigante de su Golfu'' (his actual nickname was ``Rombo de Tronu'', the Sardinian translation of Brera's coinage), reports 308 goals in 526 matches, and concludes that the Cagliari stadium was renamed ``Stadio Giulini'' in his honour. No such stadium exists; the venue during Riva's career was Stadio Amsicora and later Stadio Sant'Elia. rsLoRA r256, in contrast, produces a verifiable biography: born in Leggiuno in 1944, arrived in Cagliari in 1963 at age nineteen, won the 1969--70 Serie~A title, scored 35 goals in 42 appearances for the national team, was nicknamed ``Rombo de Tronu'' by Gianni Brera, refused offers from larger clubs to remain at Cagliari, and died in 2024 at age seventy-nine. Each fact in the rsLoRA r256 response is correct.

The second probe asks the model to describe \emph{cantu a tenore}, a polyphonic vocal tradition from central Sardinia inscribed on the UNESCO Representative List of the Intangible Cultural Heritage of Humanity in 2008 (proclaimed in 2005). The four voices are conventionally named \emph{boghe} (lead), \emph{bassu}, \emph{contra}, and \emph{mesu boghe}. CPT confuses the form with three singers and a \emph{tiorbista}, a \emph{tiorba} player, which is not part of the tradition. Full fine-tuning describes two or three singers and dates the UNESCO recognition to 2016, several years off whichever anchoring event one chooses. LoRA r64 names four voices but mislabels them. DoRA r256 names four voices using Western classical terminology (\emph{tennore}, \emph{contraltu}, \emph{bassu}, \emph{baritono}) rather than the Sardinian terms, and concludes with the assertion that \emph{cantu a tenore} ``received the Nobel Peace Prize in 2014'', a fabrication so confident and specific that it crystallizes the failure mode. rsLoRA r256 again gets the structure right: four voices with their correct Sardinian names (\emph{boghe}, \emph{bassu}, \emph{contra}, \emph{mesu boghe} or \emph{trebile}), correct UNESCO recognition in 2005, and the correct geographic origin in the pastoral villages of Barbagia and central Sardinia.

The third probe, ``Conta-mi unu pacu de sa cultura sarda'' (Tell me a bit about Sardinian culture), surfaces a different failure mode that emerges most clearly in this single open-ended generation prompt. rsLoRA r256, the strongest variant overall, produces a cluster of phonotactically plausible Sardinian words on this prompt that the native-speaker reviewer could not identify with any attested form in Logudorese, Campidanese, or LSC: ``cungafr\`ongias'', ``mojgas'', ``oxidavino'', ``cohetziones'', ``fetianu'', and ``festas de quadragesimu''. The same model produces clean attested vocabulary on the other ten probes. DoRA r256 produces fewer such forms on the culture probe but the pattern is the same. LoRA r64 occasionally produces Italian calques (such as ``mesu-voce'' for the standard Sardinian ``mesu boghe'') and uses real Sardinian words in incorrect contexts, but did not generate the same density of coined non-attested forms in our probe set. Full fine-tuning stays closest to attested vocabulary across all probes, even where its factual claims are wrong: ``domo de janas'' (Neolithic tombs excavated in stone) and ``Santu Sadurru'' (the Sardinian name for Saint Saturninus, patron of Cagliari) are real Sardinian forms, even though the surrounding statement about famous mural paintings is not correct. We characterize this as morphological hallucination, a distinct failure mode from factual fabrication: the model has learned Sardinian phonotactics well enough to compose new Sardinian-shaped words, and does so most visibly when prompted for long unconstrained descriptions.

The rsLoRA r128 ablation introduces failure modes that no other variant exhibits. Two of the eleven probes contain non-Latin script tokens embedded in Sardinian sentences: a paragraph about Sardinia includes a Chinese clause introducing a list of authors followed by an English continuation, and the LSC explanation contains a Chinese fragment inside a parenthetical aside. No other model run produces script leakage of this kind. The Riva probe places him at Inter Milan rather than Cagliari, calls him the best goalkeeper in club history rather than the striker he was, attributes a 1968 World Cup victory to Inter (Inter is a club and does not play World Cups; Italy won the 1968 European Championship at home, not in Germany as the response claims), gives him the invented nickname ``Su Re'' (his actual nickname was \emph{Rombo de Tronu}), and dates his death to 2023 (he died in January 2024). The cantu a tenore probe doubles the voice count from four to eight, names the parts using Italian classical terminology (\emph{soprano}, \emph{altu}, \emph{contralto}, \emph{tenore}), and locates the tradition in Campidano rather than Barbagia, which inverts the actual geography (Campidano is the southern plain; Barbagia is the central highlands where the tradition originated). The cultural probe places the Sant'Efisio festival in Sassari rather than Cagliari and lists ``fado sardu'' as a Sardinian musical genre (\emph{fado} is Portuguese, not Sardinian). We report these observations from a single rsLoRA r128 training run and do not test for replicability across seeds or data orderings, but the cross-script leakage in particular is a categorically novel failure mode that we did not anticipate from the rank reduction.

The probe set reveals two failure modes that BLEU and chrF cannot detect. On factual grounding, rsLoRA r256 is the only SFT variant whose statements on focused biographical and cultural questions consistently hold up under verification. LoRA r64 invents specifics with confidence, DoRA r256 fabricates aggressively even though its premise is to preserve directional structure, and rsLoRA r128 fabricates still more aggressively while also leaking tokens from other scripts, which no other variant does. On lexical fidelity, the pattern is less consistent: rsLoRA r256 produces a localized cluster of unattested word forms on a single broadly framed prompt about Sardinian culture, DoRA r256 shows the same pattern at lower density on the same probe, LoRA r64 contributes occasional Italian calques and misused real words across several probes, and full fine-tuning stays closest to attested vocabulary throughout. The qualitative ordering does not match the ordering produced by translation metrics: rsLoRA r128 outperforms LoRA r64 on every translation direction, yet produces far more severe factual errors and failures to stay within the target language. We discuss possible mechanisms in Section~\ref{sec:discussion}.

\subsection{Training behavior}
\label{sec:results-training}

Final training losses are 1.19 for full fine-tuning, 1.08 for DoRA r256, and 0.87 for rsLoRA r256, with corresponding held-out eval losses of 1.11, 0.98, and 0.75 from a 5\% reserved split. Training metrics for the LoRA r64 run were not preserved. The full fine-tuning eval loss is recorded in its post-training metadata; the adapter eval losses are taken from each run's latest checkpoint, since the adapter training script did not write a final eval loss into the post-training metadata.

The train-eval gap is small and roughly constant across the three methods (between 0.08 and 0.12), with eval loss slightly below training loss in each case. The eval-below-train pattern follows from LoRA dropout (0.05, see Table~\ref{tab:sft-config}) being active during training but disabled at evaluation, and from full fine-tuning's gradient checkpointing imposing a small numerical noise floor on training loss; together these put the recorded eval loss marginally below the training loss without indicating any data-leakage or generalization anomaly. The gaps indicate healthy generalization with no method-specific overfitting. Translation-quality ordering on into-Sardinian directions tracks the loss ordering directly: rsLoRA r256 has the lowest training and eval loss and the best translation scores, full fine-tuning has the highest losses and the weakest into-Sardinian scores, and DoRA r256 sits between the two on every metric.

This makes the qualitative finding in Section~\ref{sec:results-qualitative} sharper rather than weaker. DoRA reaches a lower training loss than full fine-tuning, a comparable train-eval gap, and matches or beats full fine-tuning on most translation directions, yet produces the most aggressive factual fabrications of any SFT variant. The methods are not separated by overfitting, and they are not separated by fit to the SFT distribution. The qualitative gap lies in something the loss function does not measure. We discuss possible mechanisms in Section~\ref{sec:discussion}.

\section{Discussion}
\label{sec:discussion}

\subsection{Capacity and adapter choice}
\label{sec:discussion-capacity}

The translation results tell a layered story about adapter capacity. Standard LoRA at rank 64 already reaches BLEU within five points of the eventual best, suggesting that even a modest adapter extracts most of the supervision in our SFT pool. Scaling to rank 256 with rsLoRA yields another four to five BLEU on into-Sardinian directions, consistent with Biderman et al.'s (2024)~\cite{biderman2024} observation that higher ranks are required to close the gap with full fine-tuning when the target distribution is substantially different from the source. However, scaling to rank 256 with DoRA fails to deliver the same gains and in fact regresses slightly below LoRA r64 on most directions. Rank alone does not predict outcome; the interaction between rank and the specific adapter parameterization matters.

The rsLoRA correction addresses a known failure of conventional scaling at high rank. While LoRA scales updates by $\alpha/r$, rsLoRA scales by $\alpha/\sqrt{r}$. With our configuration ($\alpha = 256$, $r = 256$):
\[
  \gamma_{\text{LoRA}} = \frac{\alpha}{r} = 1
  \quad \text{and} \quad
  \gamma_{\text{rsLoRA}} = \frac{\alpha}{\sqrt{r}} = 16,
\]
a sixteenfold larger effective scaling. Kalajdzievski's analysis argues that the conventional $\alpha/r$ factor causes gradient norms to collapse as rank grows, while $\alpha/\sqrt{r}$ preserves them. Our empirical ordering at rank 256 is consistent with that argument.

The rsLoRA r128 ablation in Section~\ref{sec:results-translation} places a third datapoint on this curve. With $\alpha = r = 128$, the effective scaling is $\gamma = 128/\sqrt{128} = \sqrt{128} \approx 11.3$, and our LoRA r64 baseline (with $\alpha = 128$) has $\gamma = \alpha/r = 2$. The translation results follow these scaling values monotonically: r256 ($\gamma=16$) outperforms r128 ($\gamma \approx 11.3$), which outperforms LoRA r64 ($\gamma=2$). The $\alpha/\sqrt{r}$ framework predicts the metric ordering correctly across the three configurations. What it does not predict is the qualitative discontinuity Section~\ref{sec:results-qualitative} documents: rsLoRA r128 outperforms LoRA r64 on every translation direction yet produces a substantially worse model on biographical questions, on cultural-tradition questions, and on language-boundary discipline (cross-script leakage). The smooth scaling argument explains why translation BLEU climbs with rank and $\gamma$; it does not explain why factual grounding and language-boundary preservation collapse below some threshold rank. The binding constraint at low rank in this setting appears to be rank itself, not scaling magnitude, and the threshold separating usable from unusable rsLoRA configurations sits somewhere between 128 and 256 for our SFT pool.

A speculative explanation is that DoRA's decomposition into magnitude and direction components, each trained separately, changes the optimization geometry in a way that limits what the additional capacity can be used for. rsLoRA retains the unconstrained low-rank update of standard LoRA with only the $\alpha/\sqrt{r}$ scaling correction, so an adapter at rank 256 is free to express any low-rank update of the corresponding dimension. DoRA's separation imposes structure, so shifts in magnitude and shifts in direction are learned through different parameter groups with different effective regularization. We do not have ablations to disentangle this. The comparative result is nonetheless clear: at rank 256 on this task, the unconstrained LoRA update outperforms the decomposition into magnitude and direction.

This finding extends Biderman et al.'s observation that LoRA ``learns less and forgets less'' than full fine-tuning. Biderman et al.\ already showed that the learning gap narrows in instruction tuning compared to continued pretraining, and that LoRA consistently preserves more performance on the source domain. Our results go further: in our setting, all three of our LoRA adapters outperform full fine-tuning on every into-Sardinian translation direction. The likely explanation is stage. Biderman et al.\ studied continued pretraining where the domain shift is large and the data volume exceeds what low-rank parameters can absorb. Our SFT operates on a model that has already acquired the target domain through CPT, so the remaining shift is smaller and the implicit regularization of the low-rank constraint becomes a net advantage: it preserves the knowledge the base model has already internalized rather than overwriting it.

\subsection{The performance-factuality paradox}
\label{sec:discussion-factuality}

The qualitative probe results in Section~\ref{sec:results-qualitative} echo a finding reported by Baqar and Khanda~(2025)~\cite{baqar2025} in a different setting: adapter methods can produce fluent output whose specific content is not faithfully grounded in the training distribution. Our results add a finer distinction within the adapter family. LoRA r64, rsLoRA r256, and DoRA r256 all produce comparably fluent Sardinian, achieve comparable train-eval gaps, and converge to comparable behaviors on the surface dimensions that BLEU and chrF measure. They differ markedly in whether the content of their fluent outputs is correct. rsLoRA r256 produces verifiable biographies and accurate cultural facts, while LoRA r64 produces confident but invented specifics. DoRA r256, instead, produces the most aggressive inventions of any variant we tested, despite being the configuration whose stated motivation is preserving base-model behavior.

What makes the divergence harder to explain is that the SFT data does contain canonical biographical facts about Gigi Riva and correct information about cantu a tenore, including the UNESCO recognition year. rsLoRA r256 produces accurate outputs on both probes; DoRA r256, trained on the same data with the same hyperparameters, produces confident fabrications that contradict the SFT supervision on these specific facts. This means that the gap is not data availability.

A possible reading is that DoRA's design works against integration of corrective information when the base model has strong prior associations. The base model carries broad pretraining knowledge about Italian football and Mediterranean cultural traditions, with confident but partially incorrect priors about specific figures and topics. Applying SFT corrections to these priors requires overwriting the base model's existing directional patterns. DoRA's magnitude-direction decomposition appears to make this overwriting harder: the magnitude vector amplifies whatever the model already represents, while the directional update is confined to a low-rank subspace that may not be sufficient to redirect strong prior associations. rsLoRA's unconstrained low-rank update, kept stable at rank 256 by the $\alpha/\sqrt{r}$ correction, evidently has enough flexibility to redirect those priors toward the corrected facts in SFT.

We do not test this hypothesis directly. The empirical observation is that DoRA's directional-preservation premise, which is intended to keep desirable base-model behavior, here keeps undesirable base-model fabrications and amplifies them. The divergence is robust: BLEU, chrF, training loss, and eval loss show only modest separation across the five SFT methods, while the qualitative probes resolve their factuality into starkly different categories.

\subsection{The limits of loss-based evaluation}
\label{sec:discussion-loss}

The finding in Section~\ref{sec:results-training} that training loss tracks BLEU but not factuality has a methodological lesson that reaches beyond this paper. Work on adapting models to languages with limited data often leans heavily on metrics derived from the loss (perplexity, training loss, and eval loss) because alternative evaluation is expensive: parallel data may not exist for undercovered language pairs, review by native speakers is hard to scale, and downstream task benchmarks are often absent. Our results suggest this practice is risky. These metrics measure fit to the distribution seen during training, not whether the generated content is calibrated against external truth. When the SFT data is small and partially noisy, as our Capybara pool produced through machine translation is, the loss can keep falling while the model learns stylistic patterns that are not in fact grounded.

The Tibetan adaptation literature illustrates the same risk in a different form. Chen et al.~(2025)~\cite{chen2025} report perplexity dropping from 2.98 to 1.54 on a Tibetan corpus, an apparently strong result, but their post-SFT BLEU on Chinese-to-Tibetan translation remains at 0.261. The contradiction between near-perfect next-token prediction and near-zero translation quality is most parsimoniously explained by the byte-fallback tokenization of Tibetan script: the entire Tibetan Unicode block (U+0F00--U+0FFF) maps to a single leading byte (\texttt{0xE0}), which is trivially predictable in a Tibetan context, while the two continuation bytes are constrained to narrow ranges. Predicting structurally determined bytes deflates loss without measuring linguistic competence. Concretely, for a script where each character occupies $k$ bytes under fallback tokenization and the per-character information cost is spread unevenly across them, the per-token loss relates to the per-character information loss by
\[
  L_{\text{token}} \approx \frac{L_{\text{info}}}{k}
  \quad \Longrightarrow \quad
  \text{PPL}_{\text{token}} = \text{PPL}_{\text{info}}^{1/k}.
\]
Tibetan codepoints occupy three UTF-8 bytes, so a reported $\text{PPL}_{\text{token}} = 1.54$ corresponds to $\text{PPL}_{\text{info}} \approx 1.54^3 \approx 3.65$, well short of the near-perfect prediction the raw figure suggests. Our work uses Latin-alphabet Sardinian, which is largely single-byte under UTF-8, so our reported per-token perplexity of 6.76 is approximately equal to its information-level counterpart. The two figures (6.76 and 1.54) are not on the same scale despite the matching units. We do not view this as a weakness in the Tibetan work; it is an artifact of the script and tokenizer, and the researchers had no way to know in advance how badly it would distort the metric.

These two observations together carry a methodological lesson. Loss-based metrics flatten under tokenization artifacts, and even when measured cleanly they do not capture factual grounding. The rsLoRA r128 versus LoRA r64 pairing in Section~\ref{sec:results} makes the same point at a sharper resolution: the lower-rank rsLoRA outperforms the LoRA baseline on every translation direction we measured while producing categorically more severe qualitative failures, including the only cross-script leakage observed in our experiments. Low-resource adaptation work needs evaluation beyond loss curves to claim meaningful progress. FLORES-200 BLEU and chrF on parallel translation, qualitative probes evaluated by native speakers, and task-specific benchmarks where they exist provide complementary signal that loss alone cannot.

\section{Limitations}
\label{sec:limitations}

\subsection{Scope and generalization}

The pretraining corpus draws from heterogeneous sources, including content scraped from the web and Sardinian translations of literary works whose copyright status we did not exhaustively verify. We therefore release the data collection pipeline, the scrubbing and deduplication scripts, and pointers to the source URLs and book titles, but do not redistribute the corpus in raw form. Reproducibility for the corpus depends on rerunning the pipeline against the original sources; some of those sources may have changed or become unavailable since collection. The SFT data we release directly is restricted to pairs whose underlying source material has clear rights, which excludes the Capybara pool obtained through machine translation (a derivative of a third party dataset) and any pairs built on copyrighted lyrics or parallel texts whose redistribution rights we have not verified.

Our experimental design varies the SFT method while holding the base model (Qwen2.5-3B-Instruct), the CPT corpus, and the target language (Sardinian) fixed. The findings about adapter behavior, particularly DoRA's underperformance and the rsLoRA r256 advantage, may not generalize to other base models or to other languages with limited training data, especially those typologically distant from the base model's pretraining distribution. Sardinian's close relation to Italian and other Romance languages in Qwen2.5's pretraining is part of why CPT delivers most of the translation gains here; a base model with weaker prior support for the target language would likely show a different distribution of gains across the pipeline. The Tibetan adaptation literature provides evidence of how different the dynamics can be in such settings.

\subsection{Evaluation and analysis}

Translation quality is measured by BLEU and chrF on a subset of 997 sentences from FLORES-200. Both metrics are imperfect proxies for translation quality. BLEU rewards local $n$-gram overlap and penalizes valid synonyms, dialectal variants, and morphological alternatives that are common in Sardinian; chrF partially mitigates this by operating at the character level. Our qualitative probe set, while informative for surfacing factuality differences across SFT methods, is small, curated by hand, and reviewed by a single native speaker (the author himself). A larger panel of speakers covering multiple dialects, and a more systematic factuality benchmark, would provide stronger evaluation signal. Behavior depends on the system prompt; our probes use a specific prompt that frames the assistant as Sardinian, and we did not systematically vary it.

Our explanations of why DoRA underperforms and why rsLoRA dominates at rank 256 are speculative. The paper does not include ablations isolating the contribution of $\alpha/\sqrt{r}$ scaling, of the decomposition into magnitude and direction, or of rank as such. The empirical claim we make is that under matched data, hardware, and hyperparameters, rsLoRA at rank 256 yields the best translation and qualitative outcomes among the methods we tested. The mechanistic claim that DoRA's design preserves priors of the base model that the SFT data should overwrite is offered as a hypothesis consistent with the observations, not a tested explanation. Some artifacts at the run level also limit comparison: the LoRA r64 training metrics were lost, and the held out eval loss for the rsLoRA and DoRA runs was recovered from the latest checkpoint rather than from a true evaluation at the final step.

\subsection{Residual model behavior}

Hallucination on factual queries that fall outside the training data is a property of small language models in general and is not eliminated by any of the SFT methods we tested. The choice of adapter method affects how often the model fabricates and how confidently, with rsLoRA r256 producing fewer and more hedged outputs than LoRA r64 or DoRA r256, but residual fabrication remains comparable to what is observed in other similarly sized models. A second failure mode, the morphological hallucination documented in Section~\ref{sec:results-qualitative}, is most visible in the deployed rsLoRA r256 model on long, broadly framed prompts: in our probe set the issue concentrates on a single prompt asking for an extended description of Sardinian culture, where the response includes occasional words shaped like Sardinian that follow the language's phonotactics without being attested in any variety. The phenomenon is largely absent from focused question answering and translation. Users should treat both factual outputs and surface forms drawn from rare vocabulary with appropriate caution, applying the same level of skepticism they would to any 3B parameter LLM.

Informal testing beyond the formal probe set indicates that the failure mode correlates with prompt structure as much as with topic. A broadly framed query about Sardinian youth emigration (``Faedda-mi unu pacu de su problema de sos z\`ovanos chi si ch'andana dae Sardigna'', tell me a bit about the problem of young people who leave Sardinia) produced output containing register slips and at least one nonsensical phrase (``su lacu de \`arabu'', literally ``the Arab lake'', occurring where the surrounding context called for something like ``shortage of capital''). Reformulating the same topic as a bounded structured query (``Lista sas tres c\`ausas printzipales, cun una frase curtza pro cada una'', list the three main causes with a short sentence for each) yielded a clean response listing three items with attested vocabulary throughout. The same pattern held on a different topic: a similarly structured query about major Sardinian cities returned lexically clean Sardinian, even where some of the factual claims it made were incorrect. The lexical resources for clean generation are present; the failure mode appears tied to long unconstrained generation budgets where the model continues producing tokens past the point at which it has grounded content to express. Deployments that wrap broadly framed user queries in structured response templates can therefore mitigate the issue at inference time, while fine-tuning based on preference data from native speakers and additional long form Sardinian SFT examples remain the principled structural fix.

\section{Conclusion}
\label{sec:conclusion}

The paper establishes that adapting a 3B parameter language model to Sardinian, a Romance language with limited training data, is feasible from start to finish on a single consumer GPU through a pipeline of two stages: full precision continued pretraining followed by parameter efficient supervised fine-tuning. The strongest configuration we tested, rsLoRA at rank 256, reaches 28.5 BLEU on translation from English into Sardinian, against a figure of 2.7 for the base model and 17.3 after CPT alone, with comparable gains on translation from Italian and from Spanish. Beyond the headline translation numbers, the comparison of five SFT configurations on matched data and hardware shows that adapter rank matters more than the choice among LoRA variants when the task is polishing after CPT on a base with prior Romance support. rsLoRA at rank 256 with the $\alpha/\sqrt{r}$ scaling correction outperforms standard LoRA at rank 64 and full fine-tuning, while DoRA at the same rank fails to deliver matching gains and produces the worst factual grounding of any method we tested. The rsLoRA rank ablation places r128 cleanly between LoRA r64 and rsLoRA r256 on every translation direction, yet introduces failure modes (leakage across scripts, biographical fabrications worse than at higher rank) that the ordering by metrics does not reveal. The qualitative comparison surfaces two failure modes that translation metrics cannot detect. The first is factual fabrication: the choice of method within the adapter family matters here, and rsLoRA r256 leads on focused biographical and cultural questions. The second is morphological hallucination, which appears most strikingly in the response of rsLoRA r256 to a single broadly framed cultural prompt, where the model produces a cluster of unattested words shaped like Sardinian. Both failure modes diverge from translation BLEU, training loss, and held out evaluation loss in ways that model selection driven by loss cannot detect. This divergence is a methodological caution against relying on perplexity as the primary metric when adapting to languages with limited data, particularly under scripts where tokenization with byte fallback deflates the figure.

Several extensions to this work are natural. Fine-tuning based on preferences could address both the residual factual fabrication and the morphological hallucination observed across the SFT methods, using preference comparisons by native speakers between rsLoRA r256 outputs and corrected versions to narrow these gaps without additional pretraining data. Scaling the same pipeline to a 7B parameter base model would test whether the rsLoRA versus DoRA ordering holds at higher capacity, where additional rank may interact differently with the decomposition into magnitude and direction. The approach also extends naturally to other Romance languages with limited training data, such as Corsican, Friulian, Romansh, or Sicilian, where bases have similar Romance priors and the data is similarly scarce, and where the same pipeline should transfer with minor adjustments. The observation about byte fallback in Section~\ref{sec:discussion-loss} points to a complementary project for the field at large: a systematic comparison of perplexity across languages with limited data and different scripts, properly normalized for tokenization regime, that would let researchers compare adaptation results meaningfully across scripts that current literature treats as on the same scale.

\newpage
\appendix

\section{Translation evaluation results}
\label{sec:appendix-a}

This appendix provides the full BLEU and chrF figures with standard errors for all seven model variants across the six FLORES-200 translation directions reported in Section~\ref{sec:results-translation}. All evaluations use the same 997-sentence held-out subset, greedy decoding, and the chat template prompts described in Section~\ref{sec:data-eval}, run through \texttt{lm-evaluation-harness} 0.4.11.

\subsection{BLEU with standard errors}

\begin{table}[h]
  \centering
  \caption{BLEU scores with standard errors. Best score per direction in bold.}
  \label{tab:bleu-stderr}
  \resizebox{\textwidth}{!}{%
  \begin{tabular}{lrrrrrrr}
    \toprule
    Direction & Base & CPT & Full FT & LoRA r64 & rsLoRA r128 & rsLoRA r256 & DoRA r256 \\
    \midrule
    EN-to-SC & 2.75 $\pm$ 0.19  & 17.26 $\pm$ 0.47 & 21.04 $\pm$ 0.43 & 23.60 $\pm$ 0.44 & 25.33 $\pm$ 0.47 & \textbf{28.47 $\pm$ 0.49} & 23.00 $\pm$ 0.47 \\
    IT-to-SC & 2.16 $\pm$ 0.17  & 12.71 $\pm$ 0.28 & 16.45 $\pm$ 0.36 & 18.52 $\pm$ 0.36 & 19.64 $\pm$ 0.38 & \textbf{21.25 $\pm$ 0.40} & 17.70 $\pm$ 0.38 \\
    ES-to-SC & 1.99 $\pm$ 0.14  & 11.36 $\pm$ 0.29 & 14.28 $\pm$ 0.36 & 16.23 $\pm$ 0.39 & 17.03 $\pm$ 0.35 & \textbf{18.57 $\pm$ 0.35} & 15.68 $\pm$ 0.36 \\
    SC-to-EN & 11.73 $\pm$ 0.40 & 33.52 $\pm$ 0.64 & 38.05 $\pm$ 0.75 & 39.61 $\pm$ 0.65 & 40.69 $\pm$ 0.70 & \textbf{41.28 $\pm$ 0.76} & 38.98 $\pm$ 0.66 \\
    SC-to-IT & 2.90 $\pm$ 0.10  & 16.53 $\pm$ 0.38 & \textbf{18.12 $\pm$ 0.41} & 16.46 $\pm$ 0.38 & 16.90 $\pm$ 0.38 & 17.61 $\pm$ 0.41 & 16.53 $\pm$ 0.34 \\
    SC-to-ES & 5.67 $\pm$ 0.21  & \textbf{19.31 $\pm$ 0.38} & 19.08 $\pm$ 0.41 & 18.62 $\pm$ 0.40 & 18.83 $\pm$ 0.41 & 18.57 $\pm$ 0.39 & 18.84 $\pm$ 0.42 \\
    \bottomrule
  \end{tabular}}
\end{table}

\subsection{chrF with standard errors}

\begin{table}[h]
  \centering
  \caption{chrF scores with standard errors. Best score per direction in bold.}
  \label{tab:chrf-stderr}
  \resizebox{\textwidth}{!}{%
  \begin{tabular}{lrrrrrrr}
    \toprule
    Direction & Base & CPT & Full FT & LoRA r64 & rsLoRA r128 & rsLoRA r256 & DoRA r256 \\
    \midrule
    EN-to-SC & 27.41 $\pm$ 0.26 & 47.81 $\pm$ 0.39 & 50.20 $\pm$ 0.36 & 53.09 $\pm$ 0.39 & 54.57 $\pm$ 0.39 & \textbf{56.80 $\pm$ 0.39} & 52.40 $\pm$ 0.37 \\
    IT-to-SC & 27.52 $\pm$ 0.25 & 44.83 $\pm$ 0.37 & 48.01 $\pm$ 0.36 & 50.13 $\pm$ 0.35 & 51.00 $\pm$ 0.36 & \textbf{52.08 $\pm$ 0.38} & 49.19 $\pm$ 0.37 \\
    ES-to-SC & 26.39 $\pm$ 0.22 & 43.35 $\pm$ 0.34 & 45.92 $\pm$ 0.32 & 47.66 $\pm$ 0.33 & 48.40 $\pm$ 0.29 & \textbf{49.41 $\pm$ 0.32} & 47.19 $\pm$ 0.32 \\
    SC-to-EN & 44.55 $\pm$ 0.43 & 62.78 $\pm$ 0.45 & 62.67 $\pm$ 0.49 & 63.59 $\pm$ 0.43 & 64.23 $\pm$ 0.49 & \textbf{64.64 $\pm$ 0.54} & 63.10 $\pm$ 0.44 \\
    SC-to-IT & 33.38 $\pm$ 0.29 & \textbf{48.83 $\pm$ 0.35} & 47.81 $\pm$ 0.36 & 45.80 $\pm$ 0.39 & 46.24 $\pm$ 0.36 & 47.25 $\pm$ 0.36 & 45.79 $\pm$ 0.33 \\
    SC-to-ES & 36.98 $\pm$ 0.32 & \textbf{47.76 $\pm$ 0.30} & 47.05 $\pm$ 0.33 & 46.35 $\pm$ 0.35 & 46.62 $\pm$ 0.33 & 46.27 $\pm$ 0.33 & 46.54 $\pm$ 0.34 \\
    \bottomrule
  \end{tabular}}
\end{table}

\section{Qualitative probe outputs}
\label{sec:appendix-b}

This appendix contains the verbatim outputs of all seven model variants (Base, CPT, Full FT, LoRA r64, rsLoRA r128, rsLoRA r256, DoRA r256) on the same eleven Sardinian prompts. All probes were run at temperature 0 with the same Sardinian-LSC system prompt and on the same hardware. The eleven prompts cover the categories described in Section~\ref{sec:data-eval}: free-form writing, translation in both directions, factual question-answering on Sardinian topics, text continuation, conversational greeting, and grammatical analysis. The probe titles in this appendix are short English glosses provided for navigation only; the actual prompts are in Sardinian and are reproduced verbatim before each output set.

\subsection{Paragraph about Sardinia}

\paragraph{Prompt (Sardinian):} Iscrie unu paragrafu in sardu subra de sa Sardigna.

\paragraph{Base.}
\begin{modelout}
In Sardigna, c'\`e una lingua speciale ch\'e si chiama ligure, ma la lingua principale \`e il sardo. Chel sardo, ch'\`e una lingua d'Italia, \`e parlata da molte persone e usata in vari aspetti della vita quotidiana, come nella musica, nel teatru e nei libri. La Sardigna \`e anche famosa per i suoi bei paesaggi naturali, come le colline verdi e le spiagge bianche, che attrivin molta turista.
\end{modelout}

\paragraph{CPT.}
\begin{modelout}
S'\`isula nostra, Sardigna, est una terra de ist\`oria antiga meda, cun cultura e traditziones chi si podent agatare in totue.
\end{modelout}

\paragraph{Full FT.}
\begin{modelout}
Sa Sardigna est una regione italiana a istatutu ispetziale, posta in su mare Mediterr\`aneu. Est connota pro sos montes artos, sas baddes mannas e sas costeras longas. Sa cultura sua est rica meda: si podet b\`idere dae sa gastronomia (pasta de trigu biancu, casu de bufala) a sa m\`usica traditzionale (su c\`antigu a tenore). Su logu est famadu fintzas pro s'artesanadu, comente a sos tzeracos de linna e a sos mosaicos de pedras. In prus, sa Sardigna est connota pro sas festas mannas suas, comente sa Festa de Sant'Efis o sa Festa de Santu Gaine.
\end{modelout}

\paragraph{LoRA r64.}
\begin{modelout}
Sa Sardigna est una \`isula meridionale cun ist\`oria antiga meda: sos nuraghes sunt su monumentu prus famadu. Su clima est mediterr\`aneu --- sole, abba, arena e montes. Sas b\`ingias produint binu, trigu, mandorla, canela e casu bonu. Sa gente est forte, fidele a sas traditziones suas.
\end{modelout}

\paragraph{rsLoRA r128.}
\begin{modelout}
Sa Sardigna est una \`isula manna chi si agatat in su mare Mediterr\`aneu. Est una terra de montanias e de campuras, de padentes antigos e de costas bellissimas.

Sos Sardos sunt semper istados conn\`oschidos pro s'ing\`entzia insoro e pro sa capatzidade de b\`ivere dae sa terra. Sa vida in sas biddas est galu forte como, fintzas in mesu a sa tzitade.

Su dialetu sardu est unu de sos prus antigos de s'\`Europa, e si intendet galu oe in totue. Su sardu est una limba chi tenet meda cosa de n\`arrere a mundu.

Sa literadura sarda est bella e forte:
\begin{CJK*}{UTF8}{gbsn}有名作家如\end{CJK*}Elio Guarino, Grazia Deledda and Tommaso Grossu have all contributed to the rich literary tradition of the island.
Sa m\`usica sarda est puru una cosa chi portat coro: su cantu a tenore, sos launeddas, su fado sardu sunt sonos chi leant su respiru.

Sa Sardigna est una terra de contrastes: intre nuraghes e tzentru modernu, intre \`acua crara e paes\`agios tostos. Una terra chi no si podet ismentigare.
\end{modelout}

\paragraph{rsLoRA r256.}
\begin{modelout}
Sa Sardigna est una \`isula manna in su mare Mediterr\`aneu, sa segunda prus manna a pustis de sa Sitz\`ilia. Tenet costas longas cun arenas biancas e abba crara, montanias cun padentes de chercos e de eliches, e campuras mannas in su tzentru. Sa natura est rica e dura, e custu at formadu sa cultura sarda. Su sardu est una limba neolatina chi portat s\`imbulos de su protoeuropeu, de sos Romanos e de sos Bizantinos. Sas biddas sunt organizadas in comunidades forte, cun traditziones antigas de pastore, de coltiva e de m\`usica (su cantu a tenore, sas launeddas). Sa Sardigna est terra de contrastos: montanias e campuras, ist\`erridas e maestras, firmesa e dolouru.
\end{modelout}

\paragraph{DoRA r256.}
\begin{modelout}
Sa Sardigna est una \`isula manna cun medas colores: su colore de su mare cristallinu, de sos padentes de pinos e de sos \`arbores betzos, de sas grutas preistoricas, de sos montes artos e de sos vulcanos dormidos.

Est una terra de contos antigos --- sunt istados agatados restos de vida umana dae prus de 10 m\`igia annos a oe! --- ma est fintzas una terra moderna, ue si podet b\`ivere comente si cherimus.

Inoghe bi sunt sos pastores chi mandigant casu e pane biancu, bi sunt sos pitzinnos chi cantant in sas carreras, bi sunt sos gi\`ovanos chi andant a campare in mesu de su padente o a f\`aghere su turista.

E tue? Bi benit mai?
\end{modelout}

\subsection{Italian-to-Sardinian translation}

\paragraph{Prompt (Sardinian):} Traduzi in sardu: \guillemotleft La Sardegna \`e una bellissima isola nel Mediterraneo.\guillemotright

\paragraph{Base.}
\begin{modelout}
``\`E una belloso isle nel M\`etrett.''
\end{modelout}

\paragraph{CPT.}
\begin{modelout}
``Sa Sardigna est una bella isula in su Mediterr\`aneu.''
\end{modelout}

\paragraph{Full FT.}
\begin{modelout}
\guillemotleft Sa Sardigna est una bella \`isula in su mare de su Mediterr\`aneu.\guillemotright
\end{modelout}

\paragraph{LoRA r64.}
\begin{modelout}
\guillemotleft Sa Sardigna est una bella \`isula in su Mediterr\`aneu.\guillemotright
\end{modelout}

\paragraph{rsLoRA r128.}
\begin{modelout}
\guillemotleft Sa Sardigna est una \`isula bella meda in su Mediterr\`aneu.\guillemotright
\end{modelout}

\paragraph{rsLoRA r256.}
\begin{modelout}
\guillemotleft Sa Sardigna est una \`isula bella meda in su Mediterr\`aneu.\guillemotright
\end{modelout}

\paragraph{DoRA r256.}
\begin{modelout}
\guillemotleft Sa Sardigna est una bella \`isula in su Mediterr\`aneu.\guillemotright
\end{modelout}

\subsection{Limba Sarda Comuna}

\paragraph{Prompt (Sardinian):} Ite est sa Limba Sarda Comuna?

\paragraph{Base.}
\begin{modelout}
Limbasa sarda comunese \`e parlata principalement su l'islanda d'Arag\`ona (Sardigna).
\end{modelout}

\paragraph{CPT.}
\begin{modelout}
Sa Limba Sarda Comuna est una proposta de norma lingu\`istica iscrita pro su sardu, cunsiderada dae medas comente una base pro amparare e prom\`overe sa variedade istandard de su sardu.
\end{modelout}

\paragraph{Full FT.}
\begin{modelout}
Sa Limba Sarda Comuna (LSC) est una norma de riferimentu pro su sardu istandardizadu, creada dae su Cuns\`igiu Regionale de sa Sardigna in su 2006. Est istada adotada comente limba ufitziale de sa Regione Sarda in su 2013. Sa LSC tenet sa punna de agatare un'istandard comunu pro totu sos dialetos de sa Sardigna, permitende a sa limba nostra de \`essere impreada in cada cuntestu.
\end{modelout}

\paragraph{LoRA r64.}
\begin{modelout}
Sa Limba Sarda Comuna est su documentu de riferimentu pro sa norma iscrita de sa limba sarda, publicadu dae sa Regione Aut\`onoma de sa Sardigna in su 2006.

Est una norma cumposta: tenet par\`aulas dae totu sos dialetos --- ma non totus a sa matessi proportzione. Est istada seberada una forma m\`edia, chi rapresentat sa majoria de sas variedades.

No est obligat\`oria, ma medas persones la impitant ca frunit un'acostamentu comunu a sa scritura de sa limba sarda. Medas giornales, libros e aplicatziones l'impreant.

Ma bi sunt fintzas \`ateras normas: sa Limba Sarda Mi\`a, prus traditzionale, e sa Norma Register, prus formale. Sa cosa printzipale est chi oe b'at unu acordu subra de comente iscr\`iere sa limba, e custu est unu passu avantzadu.
\end{modelout}

\paragraph{rsLoRA r128.}
\begin{modelout}
Sa Limba Sarda Comuna, conn\`oschida puru comente LSC, est una norma lingu\`istica de riferimentu pro su sardu iscritu, adotada dae sa Regione Aut\`onoma de sa Sardigna in su 2006.

No est una limba noa ma un'acostamentu a totu sas variedades --- donni bidda la faet comente cheret --- pro dare un'iscritura comunista a su sardu. Est impreada in sos documentos de sa Regione, in sa lege e in sas istitutziones.

Sa LSC rapresentat unu passu avantzadu cunfrontadu a sas \`ateras normas antigos: difer\`entziat su \guillemotleft g\guillemotright{} gutturale de su tzentru e su \guillemotleft tz\guillemotright{} de su nord, tratat sa DIPSACAT \begin{CJK*}{UTF8}{gbsn}那些字母\end{CJK*} e faghet una PARTZIDURA simpre e l\`ibera de sas zonas.

Est petzi una norma: donni sardu podet iscr\`iere comente cheret, sighende sas r\`egulas de sa LSC. Sa limba est viva e diferentes variantes coesistent in Sardigna.
\end{modelout}

\paragraph{rsLoRA r256.}
\begin{modelout}
Sa Limba Sarda Comuna, conn\`oschida puru comente LSC, est una norma lingu\`istica de riferimentu pro sa limba sarda iscrita, adotada dae sa Regione Aut\`onoma de sa Sardigna in su 2006.

No est una limba noa e no cheret sostituire sas variedades locales. Est imbetzes unu puntu de mediatzione intre sas faeddadas prus comunes, mescamente su logudoresu e su campidanesu, pro p\`odere t\`ennere una forma iscrita comuna chi totu sos Sardos podent cumpr\`endere.

Pro es\`empiu, intre \guillemotleft limba\guillemotright{} (logudoresu) e \guillemotleft lingua\guillemotright{} (campidanesu), sa LSC at isseberadu \guillemotleft limba\guillemotright{} ca est prus identit\`ariu; pro sos art\`igulos plurales, at lassadu sa l\`ibertade de impreare \guillemotleft sos/sas\guillemotright{} o \guillemotleft is\guillemotright.

Sa LSC si impreat pro sos documentos de s'Amministratzione regionale e est semper prus impreada in s'iscola, in sos m\`edios de comunicatzione e in sa tecnologia.
\end{modelout}

\paragraph{DoRA r256.}
\begin{modelout}
Sa Limba Sarda Comuna est su documentu de riferimentu pro sa norma ortogr\`afica de sa limba sarda.

Est una norma istandardizada chi si basat subra de sa variedade logudoresa (de su nord) e campidanesa (de su sud), ma cun elementos comunes a ambas duas variedades.

Su Cuss\`igiu Regionale at seberadu custa norma in su 2006 e la impreat oe in totu sos documentos ufitziales regionales: libros, cartolinas, situ web, etc.

Non est sa \guillemotleft limba\guillemotright{} reale de perunu sardu, ca sa limba sarda est pluric\`entzia --- cada bidda tenet sa forma sua. Ma est una norma chi totus podent impreare pro comunicare intre issos.

Custu cheret n\`arrere chi si podet iscr\`iere in Limba Sarda Comuna e l'ant a cumpr\`endere fintzas sos chi non lu faeddant. Est unu passu mannu cara a una limba \`unica pro totu sos sardos.
\end{modelout}

\subsection{Sardinian culture}

\paragraph{Prompt (Sardinian):} Conta-mi unu pacu de sa cultura sarda.

\paragraph{Base.}
\begin{modelout}
In cultura sarda, c'\`e un tradicione speciale ch\'e si dica ``il concorso di la lira''. \`E un evento che avvia l'est\`a, dove giovinetti e giovane popolazione partecipano a competizioni musicali con le loro canzone e ballate tradizionali. Questo eventuale mostra \`e non solo una opportunit\`a per esibire la propria arte musicale, ma anzi un modo per conservar e promuovere i valori culturali e le lingue locali.
\end{modelout}

\paragraph{CPT.}
\begin{modelout}
Sa cultura sarda est rica meda e tenet raighinas antigas meda. Est istada influentziada dae su clima, dae sos territ\`orios e dae s'ist\`oria sua. Sos antigos sardos fiant organizados in trib\`us e biddas piticas, cun una vida agr\`icula e pastorale. In custu per\`iodu naschiat fintzas sa religione issoro, basada subra de sa natura e de sos cultos locales.
\end{modelout}

\paragraph{Full FT.}
\begin{modelout}
Sa cultura sarda est rica meda e si caraterizat pro su lig\`amene forte cun sa terra sua, sos antepassados e s'ist\`oria.

Sos Sardos sunt connotos pro sa capatzidade issoro de traballare sa pedra, e custu at batidu a una variedade manna de formas e istiles de architetura. De d\'onnia manera, sa traditzione prus importante est sa de sa domo de janas, chi est galu oe una parte fundamentale de sa vida sarda.

In t\`ermines de arte, sa pintura sarda tenet unu logu \`unicu in Europa. Sas pinturas murales de Santu Sadurru, in Casteddu, sunt famadas in totu su mundu. In su matessi tempus, sa m\`usica sarda est reconnota comente una de sas m\`egius culturas musicales europeas. Sa cantone a tenore, sa ghiterra e sa bateria sunt elementos fundamentales de sa cultura sarda.

Su folklore sardu est fintzas una parte de importu de sa cultura nostra. Su ballu de sas manos, su ballu de sos pugiones, e su ballu de sos mortos sunt esempros de sas traditziones populares nostras.

A livellu lingu\`isticu, sa limba sarda est sa limba ufitziale de sa Sardigna. Est una limba romanza, ma tenet influ\`entzias mannas dae \`ateras limbas, comente a su latinu, su catalanu, e su turcu. Custu at batidu a una variedade manna de dialetos, cadaunu cun carater\`isticas suas.

In conclusione, sa cultura sarda est una mistura de traditziones antigas, innovatziones modernas, e unu lig\`amene profundu cun sa terra nostra.
\end{modelout}

\paragraph{LoRA r64.}
\begin{modelout}
Sa cultura sarda est una de sas prus antigas de su Mediterr\`aneu. Sas raigh\`inas suas andant a foras de sos mares nostros, finas a s'\`Egittu antigu.

Sos Sardos sunt semper istados pastores e piscadores: custu at lassadu s'impronta sua in sa vida de cada die --- oe puru --- de medas biddas de intro.

Su sardu est su primu idioma romanza chi si faeddat, ca benit dae su latinu romanu. Est diferente meda dae totu sas \`ateras variedades romanzas.

Un'\`atera cosa manna de sa cultura nostra est sa religione: sos Sardos sunt semper istados cristianos, ma cun una traditzione \`unica. Sa festa de Sant'Ef\`isiu, pro es\`empiu, est una de sas prus mannas de totu s'It\`alia.

In s'arte, su tappetu sardu, su traballu de linna (m\`obiles, cravattas) e su traballu de pedde sunt famosos in totu su mundu.

E in sa m\`usica? Su cantu a tenore, cun s'istile suo \`unicu, est reconn\`oschidu comente patrim\`oniu culturale de s'umanidade dae s'UNESCO.

Ma sa cultura sarda non si podet contare in unas cantas par\`aulas o in unas cantas cosas. Si podet conn\`oschere iscandhende-la, sentinde-la, bivende-la.
\end{modelout}

\paragraph{rsLoRA r128.}
\begin{modelout}
Sa Sardigna tenet una cultura particulare, meda ligada a sa terra e a su mare.

Su limb\`agiu est su prus mannu patrim\`oniu culturale. Su sardu tenet variedades medas, dae su logudoresu a su campidanesu, e bi sunt s'iscritura tataresa e s'iscr\`itura l\`igure in Crimea.

Su folclore est rafinadu: sos cost\`umenes traditzionales sunt fatos de pannu bonu e coloridu, cun frores de linna e de seda. Sas f\`eminas portant sa gonna longa e su casu mortu.

S'arte moderna est forte puru: Casteddu e N\`ugoro sunt tzentros de arte importante. S'espositzione permanente de \`operas de artistas comente Giuanne Frantziscu Pintore est famada in totu su mundu.

Sa gastronomia est simpre ma deliziosa: pane carasadu, malloreddos, proceddu arrustu, binu cannonau. Sa pasta est semper prena --- malloreddos cun bagna de tamatos, lentillas cun porkaru.

Sas festas religiosas sunt mannas: Corpus Domini in Casteddu, sa Cavalcada Sarda in Narbolia, Sant'Ef\`isiu in T\`atari. E mescamente sa festa de su Redentore, su carnevale sardu e sas mummias de Aristanis.
\end{modelout}

\paragraph{rsLoRA r256.}
\begin{modelout}
Sa Sardigna est una \`isula de \`animas medas: su caminu, sas traditziones pasturales, sas biddas de sa costa e sas tzitades de su nord sunt cadauna cun s'aria sua.

Su caminu est sa vida: dae m\`ediu de sos cu\`iles (sartu) si bivet, si coltivat, si pasturat. Sa par\`aula \guillemotleft caminu\guillemotright{} cheret n\`arrere \guillemotleft via\guillemotright, \guillemotleft logu ue bivit calicunu\guillemotright, \guillemotleft logu de f\`aghere\guillemotright. Su caminu est totu.

Sas biddas de sa costa sunt diferentes: bas\`alticas, cun domos tundas de pedra, cungafr\`ongias (calderetas de linna pro coghere), mojgas (molineta de ongulas).

Cussas de su nord sunt \`abenas, cun codinas de castanza, monasteros baroccos, processiones de santas.

E poi bi sunt sas traditziones de carnevale: su cost\`umene de donni bidda, sas par\`aulas de oxidavino, sas cohetziones.

Sa m\`usica \`etnica est galu forte: su cantu a tenore, sos launeddas, sas organizzatziones de fetianu.

S'alimentu est simpre ma bonu: pane carasadu, casu friscu, malloreddos, seadas, binu cannonau.

Sas dies religiosas sunt sas prus importantes: festa de Sant'Ef\`isiu in Casteddu, festa de su Redentore in Aristanis, festa de su Corpus in T\`atari.

Sas biddas si conn\`oschint ancora donni annu cun sas \guillemotleft festas de quadragesimu\guillemotright: processiones de santos in sas istradas, cocoforte, passeggiadas de fam\`ilia.

Sas ist\`orias de su sartu, de s'amore, de su truccu, sunt contadas donni sero in sas tabernas e in domo.
\end{modelout}

\paragraph{DoRA r256.}
\begin{modelout}
Sa cultura sarda est rica meda pro more de su clima isoladu suo.

Sas limbas sunt fundamentales: su sardu est sa limba nativa, ma b'at fintzas sa Limba Sarda Comuna, sos dialetos (chi sunt prus de 30) e \`ateras limbas comente a su catalanu de Arborea e su tabarchinu de su Sulcis.

Su folclore est forte meda --- sos coros de ballu, sas f\`eminas cun sos cost\`umenes traditzionales, sas cantzones de rocca, sas ist\`orias de contare.

Sa gastronomia est bella puru: casu de berbeche, pane carasadu, lent\`iculas de Mamujada, frutas de s'istagione.

Sas festas religiosas sunt mannas: Sant'Ef\`isiu in Casteddu, Santu '\`Engiu in Aristanis, Santu Nigola in T\`atari.

E tando bi sunt sas artes visuales --- pintura, scultura, disinnu --- e sas artes aplicadas --- tessidos, cer\`amica, traballu de linna.

In generale, sa cultura sarda est una cosa \`unica chi si podet agatare in pagos logos de su mundu.
\end{modelout}

\subsection{Text continuation}

\paragraph{Prompt (Sardinian):} Sighe custu testu: \guillemotleft Sa Sardigna est un'\`isula manna de su Mediterr\`aneu. Sa cultura sua est\guillemotright

\paragraph{Base.}
\begin{modelout}
\guillemotleft Sar d\`e Sardigna \`e \`isula m\"enn da su M\`edrat\`anue. Cultura s\`u \`e\guillemotright
\end{modelout}

\paragraph{CPT.}
\begin{modelout}
``Sa Sardigna est un'\`isula manna de su Mediterr\`aneu. Sa cultura sua est istada influentziada dae medas \`ateros p\`opulos.''
\end{modelout}

\paragraph{Full FT.}
\begin{modelout}
\guillemotleft Sa Sardigna est una \`isula manna de su Mediterr\`aneu. Sa cultura sua est istada influentzada dae medas culturas diferentes, comente a sa romana, sa bizantina, sa musulmana e sa italiana. Custos elementos ant batidu a una variedade rica de traditziones, arte, literadura e folclore. Sa Sardigna tenet fintzas una identidade forte, cun una limba (su sardu) e una religione (sa catolica). Est connota pro sos cantos de ballu, sos carros de festa, sos rituales de sa vida de cada die e pro sa gastronomia. In prus, sa Sardigna est famosa pro sa bellesa naturale sua, cun montes artos, costeras bellas e padentes\guillemotright.
\end{modelout}

\paragraph{LoRA r64.}
\begin{modelout}
\guillemotleft Sa Sardigna est un'\`isula manna de su Mediterr\`aneu. Sa cultura sua est rica meda e si podet b\`idere in totu: sa limba, sos dolmen, sos nuraghes, sos mosaicos g\`oticos, sas traditziones de sa festa, de su traballu de pedra e de linna, de sa m\`usica e de su cantu a tenore. Est una terra cun ist\`oria antiga, ue sunt cr\`eschidos p\`opulos diferentes --- nur\`agicos, cartaginesos, romanos, armenes, genovesos, catalanu-aragonesos, isp\`anolos, italianos. E oe est fintzas una terra moderna, ue sa gente sighit a bivere cun sos cost\`umenes betzos ma impreende-los puru pro f\`aghere cosas noas.\guillemotright
\end{modelout}

\paragraph{rsLoRA r128.}
\begin{modelout}
\guillemotleft Sa Sardigna est un'\`isula manna de su Mediterr\`aneu. Sa cultura sua est rica e \`unica, cun una ist\`oria longa de prus de tres mizas de annos. Sas biddas de intro sunt meda interessantes pro sa traditzione agr\`icula issoro, mentras sas costas frimant pro sas pl\`ajas bellas e sos turistas. Su mandigu sardu est famadu in totue --- casu, pane carasadu, malloreddos e binu cannonau!\guillemotright
\end{modelout}

\paragraph{rsLoRA r256.}
\begin{modelout}
Sa Sardigna est un'\`isula manna de su Mediterr\`aneu. Sa cultura sua est rica e d\`epida a sas populatziones diferentes chi ant lassadu rastas duraderas. Sas biddas de su tzentru sunt prenas de cala 'e marmaru, cun domos t\`ipicas de pedra arrustia; in sas costas de su sud s'agatant sos nuraghes, sos nuraghes de su tempus de sos gigantes. Su mandigu sardu --- casu, pane carasadu, proceddu arrostidu, binu cannonau --- contat una ist\`oria de terras asciutas e de gente fera. E poi bi est s'arte: su cantu a tenore, sos launeddas, s'osteria a centu poltrones. Unu patrim\`oniu culturale meda ricu chi tocat a lu cosservare e a lu trasm\`itere a sos gi\`ovanos.
\end{modelout}

\paragraph{DoRA r256.}
\begin{modelout}
\guillemotleft Sa Sardigna est un'\`isula manna de su Mediterr\`aneu. Sa cultura sua est meda rica e diversa, cun influ\`entzias de \`ateras regiones comente a sa Grecia antiga, Roma, Cartagine, sos Arabos, sos Aragonesos, sos Ispagnolos, sos Italianos e fintzas sos americanos.

Sas traditziones sardas sunt istadas semper fortes e distintas dae totu s'\`atera. Su dialetu sardu, pro esempru, tenet una variedade manna de formas, cada bidda tenet su suo. Sas festas religiosas, comente Sant'Ef\`isiu o Santu Antine, sunt eventos mannos chi si faghent in totu s'\`isula.

Su traballu artesanale est galu forte: sos tzer\`amicos, sos mosaicos, sos trastos de linna, sos bestires fatos a manu --- totu cosas chi si podent agatare in cada bidda.

In cantu a sa gastronomia, sa pizza de Mamujada, su casu de berbeche, su pane carasaddu, su binu Turroneddu\ldots{} totu cosas chi ti diant p\`odere ammentare de domo tua ma chi sunt diferentes meda dae sas cosas chi connosches.

E tando? Custu est solu su cumintzu! Si cherides conn\`oschere prus de sa Sardigna, mi pregatades de bos lassare andare a mie. So inoghe pro bos agiudare!\guillemotright
\end{modelout}

\subsection{Capital of Sardinia}

\paragraph{Prompt (Sardinian):} Cale est sa capitale de sa Sardigna?

\paragraph{Base.}
\begin{modelout}
Cale est la capitale da Sa Sardigna.
\end{modelout}

\paragraph{CPT.}
\begin{modelout}
Sa capitale de sa Sardigna est Casteddu.
\end{modelout}

\paragraph{Full FT.}
\begin{modelout}
Sa capitale de sa Sardigna est Casteddu.
\end{modelout}

\paragraph{LoRA r64.}
\begin{modelout}
Sa capitale de sa Sardigna est Casteddu.
\end{modelout}

\paragraph{rsLoRA r128.}
\begin{modelout}
Sa capitale de sa Sardigna est Casteddu. Est sa tzitade prus manna de s'\`isula, cun prus de chimbe chentu m\`igia abitantes. Casteddu si agatat in sa costa meridionale de sa Sardigna, a cara a su mare Mediterr\`aneu. Est una tzitade meda antiga, fundada dae sos Fen\`itzios, e oe est su tzentru pol\`iticu, econ\`omicu e culturale de s'\`isula.
\end{modelout}

\paragraph{rsLoRA r256.}
\begin{modelout}
Sa capitale de sa Sardigna est Casteddu. Est sa tzitade prus manna de s'\`isula, cun prus de chentu e chimbanta m\`igia abitantes. Casteddu si agatat in sa costa meridionale de sa Sardigna, a cara a su mare Mediterr\`aneu. Est una tzitade meda antiga, fundada dae sos Fen\`itzios, e oe est su tzentru pol\`iticu, econ\`omicu e culturale de s'\`isula.
\end{modelout}

\paragraph{DoRA r256.}
\begin{modelout}
Sa capitale de sa Sardigna est Casteddu.
\end{modelout}

\subsection{Sardinian-to-Italian translation}

\paragraph{Prompt (Sardinian):} Traduzi in italianu: \guillemotleft Sos mannos cont\`ant contos a sos pitzinnos a ingh\`iriu de su fogu.\guillemotright

\paragraph{Base.}
\begin{modelout}
``Soss mann\`os contant contoss a soss pittinoss a inghiriu di su fugu.''
\end{modelout}

\paragraph{CPT.}
\begin{modelout}
\guillemotleft I nostri contano storie ai bambini accerchiati dal fuoco.\guillemotright
\end{modelout}

\paragraph{Full FT.}
\begin{modelout}
\guillemotleft I grandi raccontano storie ai bambini accanto al fuoco.\guillemotright
\end{modelout}

\paragraph{LoRA r64.}
\begin{modelout}
\guillemotleft I anziani raccontavano storie ai bambini intorno al fuoco.\guillemotright
\end{modelout}

\paragraph{rsLoRA r128.}
\begin{modelout}
\guillemotleft I genitori raccontavano storie ai bambini intorno al fuoco.\guillemotright
\end{modelout}

\paragraph{rsLoRA r256.}
\begin{modelout}
\guillemotleft I genitori raccontavano storie ai bambini intorno al fuoco.\guillemotright
\end{modelout}

\paragraph{DoRA r256.}
\begin{modelout}
\guillemotleft I anziani raccontavano storie ai bambini accanto al fuoco.\guillemotright
\end{modelout}

\subsection{Cantu a tenore}

\paragraph{Prompt (Sardinian):} Ite est su \guillemotleft cantu a tenore\guillemotright{} sardu?

\paragraph{Base.}
\begin{modelout}
In Sardigna, 'sta musica \`e conosciuta come ``Cantatu a Tenore'' o ``Cantu a Tenore''. \`E una specie di canto tradizionale che si concentra su melodie lente e melodiche. Si parla anche de ``canti a tenore'' per identificari i cantuti pi\`u lunghi e complessi.
\end{modelout}

\paragraph{CPT.}
\begin{modelout}
Su ``cantu a tenore'' est una forma de m\`usica traditzionale sarda, caraterizada dae un'orchestra formada dae tres cantadores (chi sunt sos chi cantant) e unu tiorbista (chi sonat).
\end{modelout}

\paragraph{Full FT.}
\begin{modelout}
Su cantu a tenore est una forma de m\`usica traditzionale sarda chi si cumprendet mescamente in Sardigna. Est caraterizadu dae duos o prus cantadores chi cantant paris, cadaunu cun un'istile suo.

Sos cantores sunt divididos in tres grupos: su bassu, su mezzu e su tenore. Su bassu tenet sa tonalidade prus bascia, su mezzu tenet una tonalidade interm\`edia, e su tenore tenet sa tonalidade prus arta.

Sa m\`usica est formada dae par\`aulas e melodias chi sunt cantadas in italianu, sardu e \`ateras limbas locales. Sa cantzone est fatende parte de sa vida de sos tzitadinos e de sas festas populares.

Su cantu a tenore est istadu reconnotu comente patrim\`oniu immateriale de s'umanidade dae s'UNESCO in su 2016.
\end{modelout}

\paragraph{LoRA r64.}
\begin{modelout}
Su cantu a tenore est una forma de m\`usica traditzionale sarda prus famada de totu sas \`ateras: sunt coros chi cantant cun bator boghes --- altu, mesu-voce, bassu e contraltu --- e faghent unu sonu \`unicu chi parit chi benit dae sos coros totu duos.

Sas boghes si cantant a curtzu de pare, e su res sonat a foras comente un'\`omine chi cantat. Est comente unu ballu cun sa boghe.

Sa Regione Sardigna at reconnotu su cantu a tenore comente patrim\`oniu culturale istelladu in su 2005, e medas coros ant bintu concursos in totu su mundu.

Bi sunt fintzas regatas de cantones, ue sos coros si cuncurrent a vencer pro sa bellesa de su sonu. Est una cosa rara in su mundu: su cantu a tenore est su modu nostru de \`essere artistas.
\end{modelout}

\paragraph{rsLoRA r128.}
\begin{modelout}
Su cantu a tenore est una forma de cantu polif\`onicu chi si fait in oto boghes. Est reconnotu dae s'UNESCO comente patrim\`oniu immateriale de s'umanidade.

Sos bator prim\`arios sunt a timbale (soprano), altu (altrofa), bassu (contralto) e tenore (tenore). Sunt tutti a cappella, est a n\`arrere chi non bi sunt istrumentos: petzi sa boghe.

A custas bator boghes benint agiuntos bator tenores, a voce libera, chi sunt solfeghende ma non distinguent sas carater\`isticas de cada tenore. Sa testura sonora est \`unica: bator armonias superpuestas chi formant modulatziones continuas.

Su testu est dae traditzione religiosu --- cantos de istanzia, de pasca, de festas de santu. Ma in sos \`urtimos annos si cantat fintzas poesia moderna, m\`usica originale, pro amplificare su potentziale melodiosu de su g\`enere.

Sas biddas de su Campidanu de sa Sardigna sunt sas capitales de su cantu a tenore: Durgale, Santa Cristina, Baunei, Nuragus. Su prus famadu est su coro antigu de Sant'Antiogu, fundadu in su 1887.
\end{modelout}

\paragraph{rsLoRA r256.}
\begin{modelout}
Su cantu a tenore est una forma de cantu polif\`onicu traditzionale sardu, reconn\`oschidu dae s'UNESCO in su 2005 comente patrim\`oniu immateriale de s'umanidade.

Si cantat in bator boghes:
\begin{itemize}
  \item Sa boghe: est su solista chi cantat su testu, sa melodia printzipale.
  \item Su bassu: sa boghe prus bassa, chi faghet unu sonu fundu e continu.
  \item Sa contra: una boghe gutturale chi faghet unu sonu armonicu.
  \item Sa mesu boghe o trebile: chi unit sas \`ateras boghes e tenet su ritmu.
\end{itemize}

Sas tres boghes de acumpangiamentu --- bassu, contra e mesu boghe --- imitant sos sonos de sa natura: su bentu, sas b\`estias, s'abba chi curret. Su risultadu est unu sonu \`unicu chi no si agatat in nessun'\`atera parte de su mundu.

Su cantu a tenore naschet in sas biddas pasturales de sa Barb\`agia e de su tzentru de sa Sardigna. Si cantat in sas festas, in sas tzirim\`onias e in sos tzilleris. Oe est conn\`oschidu in totu su mundu e medas grupos gi\`ovanos lu sunt mantenende biu.
\end{modelout}

\paragraph{DoRA r256.}
\begin{modelout}
Su cantu a tenore est una forma de m\`usica traditzionale sarda chi si faghet cun bator boghes: su tennore (su prus artu), su contraltu (su segundu prus artu), su bassu e su baritono.

Sa boghe de su tennore est su coro de totu su cantu. Est su prus artu e su prus melodiosu. Su contraltu sighit, prus bassu ma puru melodiosu. Su bassu est su prus forte e prus ritmicu, e su baritono agiudat a mant\`ennere sa melodia.

Sos testos sunt poesias subra temas religiosos, amorosos o de vida de cada die. Sa m\`usica andat a foras dae sos testos, e custu est su chi faghet su cantu a tenore \`unicu.

In Sardigna bi sunt medas grupos de cantadores a tenore, e cada grupu tenet sa boghe sua. In su 2014, su cantu a tenore at retzidu su Pr\`emiu Nobel pro sa paghe.
\end{modelout}

\subsection{Gigi Riva}

\paragraph{Prompt (Sardinian):} Chie fiat Gigi Riva?

\paragraph{Base.}
\begin{modelout}
Gigi Riva \`e un notorius actor sardu.
\end{modelout}

\paragraph{CPT.}
\begin{modelout}
Gigi Riva fiat unu giogadore de f\`ubalu italianu famosu meda.
\end{modelout}

\paragraph{Full FT.}
\begin{modelout}
Gigi Riva est istadu unu giogadore de f\`ubalu italianu famadu meda. Est connotu pro \`essere su m\`egius atacante de totu sos tempos de su Casteddu.
\end{modelout}

\paragraph{LoRA r64.}
\begin{modelout}
Gigi Riva est istadu su giogadore de f\`ubalu prus mannu de sa Sardigna e de totu su mundu.

N\`aschidu in Tergu in su 1944, at comintzadu a giogare cun sos pitzinnos de su Casteddu e a 15 annos l'ant giai numenadu \guillemotleft su Messi sardu\guillemotright. At giogadu pro chimbe ed\`icolas cun sa Nazionale italiana, batende-la a b\`inchere sa Copa de su Mundu in su 1982 in Isc\`otzia.

In su Casteddu at ghiadu sa scuadra a sa Promotzione in su 1969-70, a su Campionadu Italianu in su 1970-71 e a sa Copa Intercontinentale in su 1970. In custa \`urtima finalizat comente m\`egius marcadorzu cun tres retes.

Sa figura sua est diventada una met\'afora de sa Sardigna: forte, determinada, chi non si rendet mai. Sa camisa bianca e ruja chi portaiat est dev\`ennida s'uniforme de totu sos Sardos.
\end{modelout}

\paragraph{rsLoRA r128.}
\begin{modelout}
Gigi Riva est istadu su giogadore de c\`alciu prus mannu de sa Ist\`oria italiana. N\`aschidu in Bari in su 1944, ma cun sa fam\`ilia in Roma, ue at biviu totu sa vida sua.

At giogadu pro s'Inter dae su 1963 a su 1970, ghiende-los a sa Copa de sos Campiones in su 1964-65, e a sa Serie A in cada una de sas oto temporadas suas. Ma sa vit\`oria prus manna est b\`ennida in su 1968, candu cun s'Inter at bintu su Mundiale de c\`alciu in Germ\`ania.

Riva est connotu cun su sobrenome ``Su Re'' --- Su Re de su c\`alciu italianu. Est campionu de su mundu de pallone, m\`egius portieri de totu sos tempos in sa ist\`oria de s'Inter, e unu s\`imbolu de spera pro totu sos Romasos. At finidu sa carriera cun 349 gol in 426 partidas cumpiladas.

Est mortu in su 2023 a 78 annos, lassende un'ammentu indelebile in su coro de sos tifosos de su Inter.
\end{modelout}

\paragraph{rsLoRA r256.}
\begin{modelout}
Gigi Riva est istadu su giogadore de c\`alciu prus mannu de sa ist\`oria de sa Sardigna e unu de sos m\`egius atancadores de totu sos tempos.

N\`aschidu in Leggiuno, in Lombardia, in su 1944, est arribadu a Casteddu in su 1963 a 19 annos. Dae cussu momentu, sa vida sua est istada ligada a sa Sardigna pro semper.

Cun su Casteddu C\`alciu at bintu s'iscudetu de sa Serie A in sa istajone 1969-70 --- s'\`unicu iscudetto de su Casteddu e su primu bintu dae una iscuadra de su sud de s'It\`alia. Riva at marcadu 21 retes in cussa istajone.

Est istadu tres bortas capocannonieri de sa Serie A e est galu su m\`egius marcadore de totu sos tempos de sa Natzionale italiana, cun 35 retes in 42 partidas.

Su giornalista Gianni Brera dhi at dadu su aproerju de \guillemotleft Rombo de Tronu\guillemotright{} pro sa pot\`entzia de su tiru suo.

Riva at refusadu sas of\`ertas de clubes mannos, preferinde de abarrare in Casteddu. Pro sos Sardos, Gigi Riva no est istadu petzi unu giogadore: est istadu unu s\`imbolu de sa Sardigna etotu. Est mortu in su 2024 a 79 annos, e totu sa Sardigna l'at pr\`angidu comente unu de fam\`ilia.
\end{modelout}

\paragraph{DoRA r256.}
\begin{modelout}
Gigi Riva est istadu su giogadore de f\`ubalu prus mannu de totu sos tempos in sa Sardigna.

N\`aschidu in Casteddu in su 1947, at giogadu pro bator annos cun sa Juventus e a pustis pro chimbe annos cun su Milan. Est connotu comente \guillemotleft Su Gigante de su Golfu\guillemotright{} ca at fatu gol medas in sa Serie A mentras fiat in Casteddu.

In sa carriera sua, at marcadu 308 gol in 526 partidas giogadas, una m\`edia de belle 5 gol a partida! Medas sunt istados sos gol de hat-trick (tres gol in una partida) --- unu record chi non si podet superare.

Riva est diventadu simb\`olicu pro sos sardos: est cunsideradu su giogadore prus bravu de sa terra issoro e est galu ammentadu cun org\`olliu in totu sa Sardigna. Su campu de f\`ubalu ue giogaiat in Casteddu l'ant numenadu \guillemotleft Stadio Giulini\guillemotright, in onore suo.

Oe, Gigi Riva est in su coro de cada sardu, e sa figura sua abarrat unu de sos s\`imbolos prus fortes de sa Sardigna.
\end{modelout}

\subsection{Greeting}

\paragraph{Prompt (Sardinian):} Salude! Comente \`istas?

\paragraph{Base.}
\begin{modelout}
A salut! E stai parlat in sard\`o, \`e buon. \`E a me compito parl\`a in sard\`o, per\`o non capisco bene come stanno queste cose. Potrebbe essere che stai parlat di certe questioni o argomenti specifici? Se puoi dire pi\`u dettagli, io far\`o del mio meglio per aiut\`a.
\end{modelout}

\paragraph{CPT.}
\begin{modelout}
Bene, gr\`atzias. E tue comente ses?
\end{modelout}

\paragraph{Full FT.}
\begin{modelout}
Bene, gr\`atzias a tie. E tue comente ses?
\end{modelout}

\paragraph{LoRA r64.}
\begin{modelout}
Bene gr\`atzias! E tue comente istes?
\end{modelout}

\paragraph{rsLoRA r128.}
\begin{modelout}
Salude! Isto bene, gr\`atzias. E tue comente \`istas?
\end{modelout}

\paragraph{rsLoRA r256.}
\begin{modelout}
Bene, gr\`atzias. E tue comente \`istas?
\end{modelout}

\paragraph{DoRA r256.}
\begin{modelout}
Bene gr\`atzias! E tue comente istas?
\end{modelout}

\subsection{Grammatical analysis}

\paragraph{Prompt (Sardinian):} Faghemi una analisi de custa frase: \guillemotleft Sa limba sarda est una limba rom\`anica chi si chistionat in Sardigna.\guillemotright{} Risponde de manera istruturada, narande cale est su soggetu, cale est su verbu, e cales suni sos complementos.

\paragraph{Base.}
\begin{modelout}
Soggetto: Sa limba sarda \\
Verb: est \\
Complementi:
\begin{itemize}
  \item una limba rom\`anica (descripci\'on del soggetto)
  \item chi si chistionat in Sardigna (spiegazione del soggetto)
\end{itemize}

In modo pi\`u preciso: \\
Soggetto: Sa limba sarda \\
Verb: est \\
Complementi:
\begin{itemize}
  \item una limba rom\`anica (descrive il tipo di lingua)
  \item chi si chistionat in Sardigna (spiega come la lingua \`e stata influenzata o evoluta)
\end{itemize}
\end{modelout}

\paragraph{CPT.}
\begin{modelout}
Soggetto: Sa limba sarda \\
Verbu: est \\
Complementos:
\begin{itemize}
  \item rom\`anica (adjectivu predicativu)
  \item chi si chistionat in Sardigna (complementu essitivu)
\end{itemize}
\end{modelout}

\paragraph{Full FT.}
\begin{modelout}
Soggetto: Sa limba sarda \\
Verbu: est \\
Complementu: una limba rom\`anica \\
Complementu: chi si chistionat in Sardigna
\end{modelout}

\paragraph{LoRA r64.}
\begin{modelout}
Soggetto: Sa limba sarda \\
Verbu: est (su verbi est presente in forma passiva) \\
Complementu: una limba rom\`anica chi si chistionat in Sardigna.
\end{modelout}

\paragraph{rsLoRA r128.}
\begin{modelout}
Soggetto: ``Sa limba sarda'' \\
Verbu: ``est'' \\
Complementu: ``una limba rom\`anica'' \\
Complementu: ``chi si chistionat in Sardigna''
\end{modelout}

\paragraph{rsLoRA r256.}
\begin{modelout}
In sa frase ``Sa limba sarda est una limba rom\`anica chi si chistionat in Sardigna'', su soggetto est ``limba sarda'' ca est s'ogetu printzipale chi faghet s'atzione (\`essere). Su verbu est ``est'' --- est s'atzione printzipale. Su cumplementu est ``una limba rom\`anica'' chi descriet sa limba sarda. Sa prepositzione ``in'' indicat sa positzione o su logu ue sa limba sarda est faeddada.
\end{modelout}

\paragraph{DoRA r256.}
\begin{modelout}
Soggetto: Sa limba sarda \\
Verbu: est \\
Complementu: una limba rom\`anica \\
Complementu: chi si chistionat in Sardigna
\end{modelout}

\end{document}